\documentclass[journal]{IEEEtran} 
\IEEEoverridecommandlockouts

\usepackage[utf8]{inputenc}

\usepackage{comment}
\usepackage{xspace}
\usepackage{graphicx}
\usepackage{overpic}
\usepackage{svg}
\usepackage{xcolor}
\usepackage{bm}
\usepackage[hang,flushmargin]{footmisc}
\usepackage{booktabs}
\usepackage{tabularx}

\usepackage{tabularx}
\usepackage{multirow}
\usepackage{colortbl}
\usepackage{float}
\usepackage{placeins}

\usepackage{array}
\newcolumntype{P}[1]{>{\centering\arraybackslash}p{#1}}  

\usepackage{etoolbox}
\makeatletter
\patchcmd{\@makecaption}
{\scshape}
{}
{}
{}
\newcommand{\onetagright}{\tagsleft@false}
\makeatother

\usepackage{amsmath}
\usepackage{amssymb}
\usepackage{amsthm}
\usepackage{siunitx}

\newtheoremstyle{main}
{1em}                                                
{1em}                                                
{\itshape}                                        
{0pt}                                                
{\scshape}                                           
{\\*}                                                
{2pt}                                                
{\thmname{#1}\thmnumber{ #2}: \thmnote{\itshape #3}} 

\usepackage{environ}

\usepackage[linesnumbered,ruled,noend]{algorithm2e}
\usepackage{algpseudocode}
\newcommand{\removelatexerror}{\let\@latex@error\@gobble}

\usepackage[
	activate   = {true},
	protrusion = false,
	expansion  = true,
	kerning    = true,
	spacing    = true,
	tracking   = false,
	auto       = true,
	selected   = true,
	factor     = 2000,
	stretch    = 50,
	shrink     = 20,
]{microtype}

\usepackage{csquotes}
\usepackage[
	maxbibnames=99,
	maxcitenames=2,
	natbib=true,
	style=numeric-comp,
	backend=biber,
	sorting=none,
	giveninits=true,
	url=false,
	doi=false,
	eprint=false,
	isbn=false,
]{biblatex}

\definecolor{gg}{RGB}{240, 74, 0}

\makeatletter
\def\@IEEEBIOskipN{1.25\baselineskip}
\makeatother

\makeatletter
\let\NAT@parse\undefined
\makeatother
\usepackage[pdfa,colorlinks,bookmarksopen,bookmarksnumbered,allcolors=gg]{hyperref}
\usepackage[english]{babel}

\usepackage[nameinlink,capitalise]{cleveref}
\crefname{line}{line}{lines}
\crefname{figure}{Fig.}{Figs.}
\Crefname{figure}{Fig.}{Figs.}
\crefname{equation}{Eq.}{Eqs.}
\Crefname{equation}{Eq.}{Eqs.}
\crefname{section}{Sec.}{Secs.}
\Crefname{section}{Sec.}{Secs.}
\crefname{definition}{Def.}{Defs.}
\Crefname{definition}{Def.}{Defs.}
\crefname{algorithm}{Alg.}{Algs.}
\Crefname{algorithm}{Alg.}{Algs.}
\crefname{assumption}{Asm.}{Asms.}
\Crefname{assumption}{Asm.}{Asms.}
\crefname{subassumption}{Asm.}{Asms.}
\Crefname{subassumption}{Asm.}{Asms.}
\Crefname{problem}{Problem}{Problems}
\crefname{problem}{Problem}{Problems}

\usepackage{flushend}

\usepackage[inline]{enumitem}
\setlist{itemsep=1pt, topsep=2pt, parsep=0pt}
\usepackage{mathtools}

\newcommand{\ompl}{\textsc{ompl}\xspace}
\newcommand{\cspace}{\mbox{\ensuremath{\mathcal{Q}}}\xspace}

\newcommand{\cgoal}{\mbox{\ensuremath{\mathcal{Q}_{goal}}}\xspace}
\newcommand{\cfree}{\mbox{\ensuremath{\mathcal{Q}_{free}}}\xspace}
\newcommand{\cstart}{\mbox{\ensuremath{q_{start}}}\xspace}
\newcommand{\workspace}{\mbox{\ensuremath{\mathcal{W}}}\xspace}

\newcommand{\dof}{\textsc{d\scalebox{.8}{o}f}\xspace}
\newcommand{\psprm}{\textsc{ps-prm}\xspace}

\newcommand{\eg}{\emph{e.g.},\xspace}

\usepackage{xcolor}
\usepackage{pifont} 

\newcommand{\cmark}{\textcolor{green!60!black}{\ding{51}}} 
\newcommand{\xmark}{\textcolor{red!70!black}{\ding{55}}}   

\title{\fontsize{17pt}{24pt}\selectfont \bf
Look as You Leap: \\
Planning Simultaneous Motion and Perception for High-\dof
Robots
}
\author{Qingxi Meng$^{1}$, Emiliano Flores$^{1}$, Carlos Quintero-Peña$^{1}$, Peizhu Qian$^{1}$, Zachary Kingston$^{1}$, \\ Shannan K. Hamlin$^{3}$, Vaibhav Unhelkar$^{2}$, and Lydia E. Kavraki$^{2}$%
\thanks{
This work was supported in part by NSF 2326390 and NSF CCF-2336612. 
(Corresponding author:  Lydia E. Kavraki, Qingxi Meng.)}%
\thanks{$^{1}$Qingxi Meng, Emiliano Flores, Carlos Quintero-Peña, Peizhu Qian, and Zachary Kingston are with the Department of Computer Science, Rice University, Houston, TX 77005 USA {\tt\footnotesize (e-mail: qm15@rice.edu)}.}%
\thanks{$^{2}$Vaibhav Unhelkar and Lydia E. Kavraki are with the Department of Computer Science, Rice University, Houston, TX 77005 USA, and also with the Ken Kennedy Institute, Rice University, Houston, TX 77005 USA {\tt\footnotesize (e-mail: vaibhav.unhelkar@rice.edu; kavraki@rice.edu)}.}%
\thanks{$^{3}$Shannan K. Hamlin is with the Houston Methodist Academic Institute, Houston, TX 77030 USA {\tt\footnotesize (e-mail: shamlin@houstonmethodist.org)}.}%
}

\begin{document}
\maketitle
\thispagestyle{empty}
\pagestyle{empty}

\begin{abstract}
Most common tasks for robots in dynamic spaces require that the environment is regularly and actively perceived. The perception task considered in this work can represent a broad range of robot perception objectives, including object detection, human activity recognition, and human face detection. For example, a service robot may need to continuously detect and localize a target object during manipulation, while an assistive robot in a healthcare setting may need to maintain reliable perception of a human face or activity for interaction and safety monitoring. These tasks impose perception constraints on the robot motion. However, solving motion and perception tasks simultaneously is challenging, as these objectives often impose conflicting requirements. Furthermore, while robots must react quickly to changes in the environment, directly evaluating the quality of perception (e.g., object detection confidence) is often expensive or infeasible at runtime. This problem is especially important in human-centered environments, such as homes and hospitals, where effective perception is essential for safe and reliable operation. In this work, we address the challenge of solving motion planning problems for high-degree-of-freedom (\dof) robots from a start to a goal configuration with continuous perception constraints under both static and dynamic environments. Our solution is a GPU-parallelized perception-score-guided probabilistic roadmap planner with a neural surrogate model (\psprm). Unlike existing active perception-, visibility-aware or learning-based planners, our work jointly considers perception tasks and constraints when searching for a solution to the motion planning problem. Our method uses a neural surrogate model to approximate perception scores, incorporates them into a roadmap-based solution, and leverages GPU parallelism to enable efficient online replanning in dynamic settings. We demonstrate that our planner, evaluated on high-\dof robots, outperforms RL- and trajectory-optimization-based baseline methods in both static and dynamic environments in both simulation and real-robot experiments.
\end{abstract}


\section{Introduction}
\label{sec:introduction}

\begin{figure}[t]
    \centering
    \includegraphics[width=1.0\columnwidth]{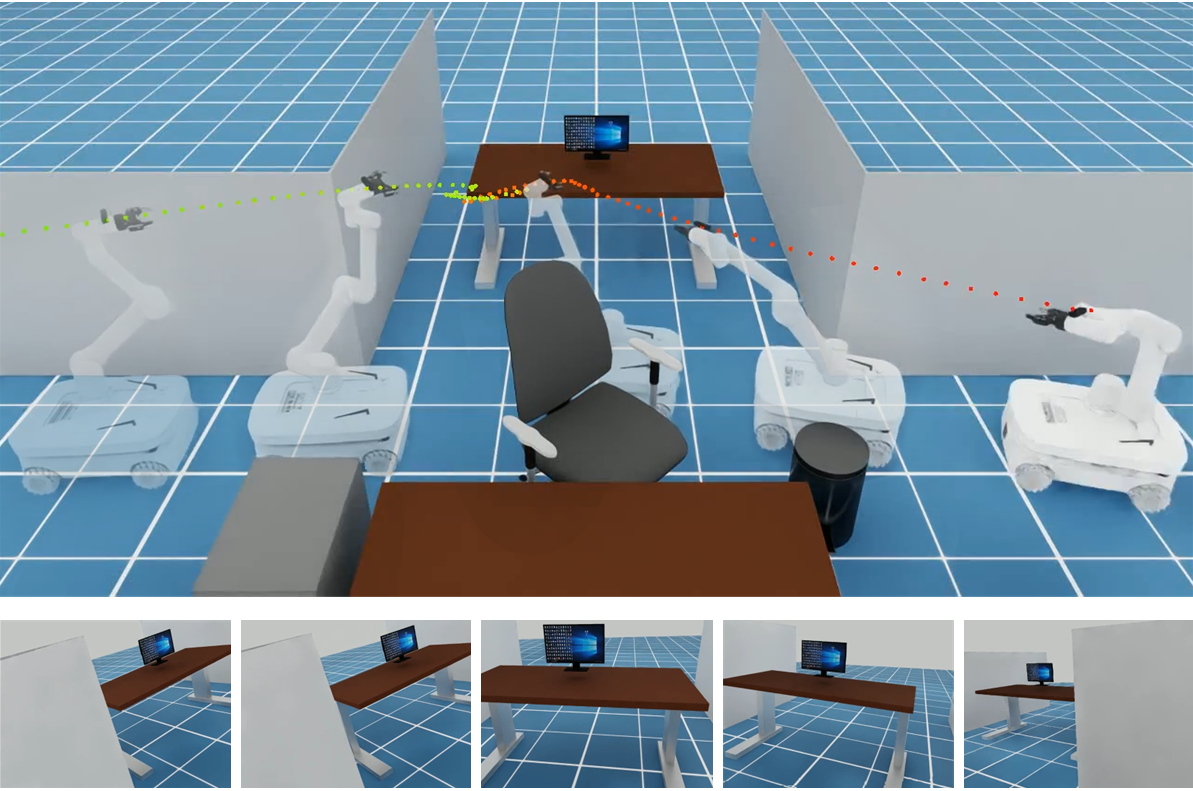}
    \caption{A UR5 on a differential-drive base moves through a cluttered environment while tracking a monitor with its wrist-mounted camera. \textbf{Top:} the robot's trajectory (fading right to left) as it maintains a high detection rate. \textbf{Bottom:} camera views for the configurations shown above.
}
\label{fig:ur5_demo}
\end{figure}

Achieving mobile manipulation (e.g., navigating to a table to grasp a tool) alongside perception tasks (e.g., tracking an object or a human) remains highly challenging. In this work, we address the problem of solving motion planning problems for high degree-of-freedom (\dof) robots from a start to a goal configuration while optimizing a given perception task in both static and dynamic environments. The perception task considered in this work can represent a broad range of perception objectives, including object detection or tracking, human activity recognition, and face detection. We define the perception quality as a task-dependent numerical measure of how well the perception task is achieved from a given continuous robot trajectory. Examples include object detection confidence scores, human pose estimation accuracy, face detection confidence, or object tracking reliability metrics. A fundamental challenge emerges: the objectives of motion and perception often conflict. The perception tasks could impose non-trivial constraints on robot motion, as performance varies with object shape, occlusion, distance, and viewing angle~\cite{yin2021center, zheng2023deep, minaee2021image}. For example, achieving reliable perception may require the robot to move closer to a target or adopt specific viewing angles, which can conflict with motion objectives such as maintaining clearance from obstacles, respecting kinematic limits, or computing/optimizing for a short and smooth path. At the same time, planning for high-degree-of-freedom (\dof) robots with perception constraints must account for the complex, nonlinear relationship between the robot's configuration and the camera pose.

As an example, consider the situation in \cref{fig:ur5_demo}. A mobile manipulator with a wrist-mounted camera must move from a start to a goal configuration through a cluttered environment while observing a monitor at the end of the hallway. The robot must reach its goal with a trajectory that avoids occlusions and achieves high tracking confidence.
\FloatBarrier

The planning challenges are further compounded in dynamic environments, where objects and people move. In such settings, robots must continuously adapt to changes while achieving high perception quality~\cite{nageli2017real, blasi2020path, piazza2024mptree, mercy2016real}.
For example, a key motivation of this work comes from human-centered environments such as nurse training settings~\cite{aloush2018nurses, quintero2023robotic}, where robotic tutors are used to support skill acquisition. In such an environment~\cite{qian20astrid}, the robot must continuously monitor a trainee to detect compliance with sterile techniques and provide timely feedback while simultaneously moving through the environment. \cref{fig:nursing_environment} and \cref{fig:real_robot_dynamic} illustrate this challenge. In these settings, the robot must continuously track nurses’ faces while navigating through a cluttered clinical environment. 
However, naively evaluating perception performance (e.g., detection confidence) for perception-aware planning is prohibitively expensive in real environments~\cite{singh2024synthetica} because it would require physically placing the sensor at a large number of distinct poses throughout the environment. This motivates the need for efficient mechanisms to deliver perception quality and support fast, responsive replanning in high-dimensional spaces. 

Existing works that consider both motion and perception are not designed for the problem examined in this paper. Informative path planners~\cite{tordesillas2022_panther,singla2021_memory,masnavi2024_differentiable,masnavi2022_visibility} aim to find paths that maximize information gain while respecting constraints such as path length or budget, but they often become computationally expensive, especially when the problem size grows. Perception-aware path planners are typically designed for low-\dof systems~\cite{falanga2018_pampc,ichter2020_perception,loquercio2021_learning,song2023_learning} or rely on overly simplified perception models~\cite{bartolomei2020_perception}, such as keeping a point of interest centered in the field of view. Active perception approaches~\cite{bajcsy1988active, aloimonos1988active, soatto2009actionable, liu2021active, davison2002simultaneous, vidal2010action, mostegel2014active, sadat2014feature, morilla2024perceptual} focus on selecting discrete viewpoints to maximize scene understanding, rather than ensuring consistent perception quality along a motion trajectory. Moreover, most of these approaches are not designed to operate in dynamic environments where objects or humans may move and where continuous adaptation is needed. As a result, they lack the responsiveness required for real-time perception-aware planning in practical settings.

In this paper, we propose \textbf{Perception}-\textbf{Score}-guided \textbf{Probabilistic Roadmap} Planning (PS-PRM): a GPU-parallelized, roadmap-based planner that explicitly incorporates the estimated quality of a given perception task (e.g., object detection confidence score) into motion planning for high-degree-of-freedom (DOF) robots. The perception score is predicted by a neural surrogate model trained to approximate perception quality for specific tasks.
A key aspect of our formulation is that perception quality is modeled in SE(3) space rather than in robot-specific configuration space. By basing the surrogate perception model on the camera pose and its relative transform to the target, it becomes robot- and environment-agnostic, enabling direct reuse across different high-\dof{} platforms with similar sensor setups. Fast forward kinematics (FK) allows this model to still be directly and efficiently linked to the configuration space and enables better approximation of continuous perception performance through interpolation between configurations. This design choice allows our planner to generalize beyond a single robot and environment, a property we demonstrate in our experiments. \psprm jointly optimizes for both motion and perception quality by associating each configuration with an estimated perception score and using these scores to guide roadmap construction and path selection. To efficiently approximate perception quality during planning, our neural surrogate model is trained on data from various objects and human targets that are typically present in the specific environment the robot is expected to operate. To account for occlusions in cluttered environments, we incorporate a ray-casting-based pipeline that dynamically adjusts perception estimates. Finally, to enable fast replanning in dynamic environments, we develop a GPU-parallelized framework that executes collision checking, FK, and perception evaluation in batch. These components allow our method to scale to high-dimensional planning problems while achieving high perception quality in both static and dynamic scenarios. 

In this work, we make the following key contributions:
\begin{itemize}
\item A new problem formulation for motion planning with high degree-of-freedom (\dof) robots that plans from a start to a goal configuration while optimizing perception quality in both static and dynamic environments.
\item A perception score estimation pipeline that uses a neural surrogate model and ray casting to efficiently approximate  perception quality while accounting for occlusions in cluttered scenes, extracting more complex and computationally expensive information than landmarks and informative regions.

\item  A planner that uses the neural surrogate model and a roadmap-based approach.

\item A batch-processing infrastructure that accelerates collision checking, FK and perception evaluation using GPU parallelism, enabling real-time replanning.
\item Extensive simulation and real-robot experiments on four different high-\dof robot systems, demonstrating that \psprm consistently improves both perception performance and planning efficiency compared to baseline methods.
\end{itemize}

\section{Problem Statement}
\label{sec:problem_statement}

We consider a robot with $k$ controllable joints and an onboard camera with a limited and known field of view. The robot's configuration space is $\cspace \subseteq \mathbb{R}^k$. The robot is located in a workspace $\workspace \subseteq \mathbb{R}^3$ that contains objects and people of interest that need to be monitored, as well as obstacles to be avoided for safety. The robot has access to the pose and geometry of all obstacles in $\workspace$ and the positions of the centroids of objects of interest. As the robot moves, the objects or humans of interest may also move, and we assume the robot has real-time access to their poses. The subset of all robot configurations that are collision-free is denoted by $\cfree \subseteq \cspace$. The motion planning problem is to find a path $\sigma: [0, 1] \rightarrow \cfree$ that assigns a collision-free configuration to each input parameter, starting from $\sigma(0) = \cstart$ and ending in a goal region $\sigma(1) \in \cgoal$.

We additionally define a motion cost function $c_m : \Sigma_{\mathcal{Q}} \rightarrow \mathbb{R}^{+}$, representing the length of a path. To consider perception quality, we define a perception score function $p_s : \Sigma_{\mathcal{Q}} \rightarrow [0, 1]$ which quantifies the performance of a perception task along $\sigma$. Based on this perception score, a perception cost is defined as $c_p=1-p_s$, with which we can define a composite cost combining motion and perception quality $c(\sigma) = c_m(\sigma) +  c_p(\sigma)$.

We denote the set of all paths in $\mathcal{Q}$ as $\Sigma_{\mathcal{Q}}$, and all collision-free paths as $\Sigma_{\mathcal{Q}_{\mathrm{free}}}$. The central problem to solve is to find a collision-free path $\sigma^* : [0,1] \rightarrow \mathcal{Q}_{\text{free}}$ that minimizes the composite cost function $c(\sigma) = c_m(\sigma) + \alpha \cdot c_p(\sigma)$, where the trade-off between motion and perception cost is governed by a scalar weight $\alpha \geq 0$. The optimal path is given by $\sigma^* = \arg\min_{\pi \in \Sigma_{\mathcal{Q}_{\text{free}}}} \left[ c_m(\sigma) + \alpha \cdot c_p(\sigma) \right]$, if such a path exists.

In our problem setting, minimizing $c_m$ implies shorter, better paths, while minimizing $c_p$ is implicitly maximizing the perception score through the path. For example, in an object tracking task, the perception score may correspond to the average tracking confidence score obtained as the robot moves along the path. Our problem formulation seeks to minimize a weighted sum of motion and perception costs, aiming to find a path that is both short and achieves good perception performance.

\section{Related Work}
\label{sec:related_work}
In this section, we discuss three categories of related problems examined in this paper: informative path planning, perception-aware path planning, and active perception. While all of these approaches consider perception and motion, they address fundamentally different problem formulations compared to ours.

\subsection{Informative Path Planning}
There have been many works
on the problem of informative or information-rich path planning ~\cite{low2008adaptive, hollinger2014sampling, ruckin2022adaptive}. The general problem 
is to maximize a metric of information quality (e.g., probability of locating a target, accuracy of the inspection, or quality of the survey) while satisfying constraints on energy, time or perception. The informative motion planning problem of maximizing information gathered subject to a constraint is particularly challenging because it typically requires searching over a large and complex space of possible trajectories. Such problems have been shown to be NP-hard~\cite{singh2009efficient}, or even PSPACE-hard~\cite{reif1979complexity}, depending on the form of the objective function and the space of possible trajectories. Several general solvers for robotic information gathering problems utilize combinatorial optimization techniques to search over a discrete grid. The recursive greedy algorithm~\cite{singh2009efficient} is one example that achieves bounded performance for submodular objective functions. Branch and bound techniques have also been proposed that only require the objective function to be monotonic~\cite{binney2012branch}. Another approach is to utilize a finite-horizon solver that solves the problem for only a portion of the budget at a time~\cite{hollinger2009efficient}. However, the resulting solvers typically require computation exponential in the size of the problem instance due to the large blowup in the search space. Thus, they are mostly shown with 2D environments and with very few obstacles. 

Another related line of work is planning under uncertainty or planning in information spaces, which focuses on selection of trajectories that minimize the localization uncertainty. This problem has generally been solved with Partially Observable Markov Decision Processes (POMDPs) or through graph-search in the belief space~\cite{bonet2000planning}. While these approaches are well-established, their computational complexity grows exponentially in the number of possible actions and observations.

To overcome these issues, the use of sampling-based motion planning algorithms, such as the RRT~\cite{lavalle00} and the probabilistic roadmap (PRM)~\cite{kavraki1996probabilistic}, has increased enormously in past years. Such algorithms have the advantage of quickly exploring a space of interest to achieve a feasible solution. Variants of sampling-based motion planning have also been proposed to maximize some metric of information quality. The prior algorithms have been shown to solve related problems for certain objective functions, including information-rich RRTs~\cite{levine2010information} using Fisher information matrices and belief-space RRTs~\cite{bry2011rapidly}. 

Unlike these works, this paper specifically focuses on selecting a trajectory that balances high perception quality with short motion length. Furthermore, these prior algorithms often focus on 2D environments~\cite{levine2010information, bry2011rapidly, hollinger2014sampling, ruckin2022adaptive} and their information metrics do not apply to the perception task we consider in this work where the approximate analytical model is infeasible.

\subsection{Perception-aware Path Planning}
Perception-aware path planning has been studied extensively for Unmanned Aerial Vehicles (UAVs)  and Micro Air Vehicles (MAVs) because a perception-aware component is especially important when flying in dynamic environments. Furthermore, a consistent detection of the moving obstacles is necessary to obtain a good estimate of their locations and prediction of their future trajectories. 

Specifically, our work relates to recent work on optimizing motion and perception tasks for UAVs and MAVs in applications such as quadrotor flying~\cite{falanga2018_pampc,ichter2020_perception,loquercio2021_learning,song2023_learning}, under dynamic environments~\cite{tordesillas2022_panther,singla2021_memory,masnavi2024_differentiable,masnavi2022_visibility} and for scenes with  semantic classes~\cite{bartolomei2020_perception}.
In most cases, the perception task aims at maintaining visibility of landmarks and informative regions for visual odometry~\cite{bartolomei2020_perception}, reducing the velocity of the projection of points of interest on the camera plane to avoid blurred motion~\cite{falanga2018_pampc,lu2022_realtime} or guaranteeing the visibility of obstacles for safety~\cite{penin2018_vision,masnavi2022_visibility,tordesillas2022_panther}. Further relevant work addresses the problem of having a UAV record or chase a target~\cite{penin2018vision, penin2017vision, thomas2017autonomous, li2021pcmpc} For example, \cite{thomas2017autonomous} focused on tracking a moving target with a downward-facing camera, while ~\cite{penin2018vision} proposed a way to follow a moving target while avoiding other static obstacles in the environment.

For UAVs, most methods formulate the problem as optimization using Model Predictive Control (MPC)~\cite{falanga2018_pampc,penin2018_vision,tordesillas2022_panther,masnavi2022_visibility} or Reinforcement Learning (RL)~\cite{singla2021_memory,loquercio2021_learning,song2023_learning,tordesillas2023_deep}.
MPC-based methods jointly optimize perception and motion for agile UAV navigation using numerical optimization, without obstacles~\cite{falanga2018_pampc} or for static obstacles and occlusions~\cite{penin2018_vision}.
For example, the work~\cite{bartolomei2020_perception} steer the UAV towards informative areas using a path planner while keeping landmarks in the camera field of view through optimization.
RL-based methods create policies that directly map from sensor measurements to controls while keeping visibility of regions of interest.
The policy learner has access to privileged information during training, \emph{e.g.,}~by using MPC~\cite{zhang2016_learning}, a planner~\cite{loquercio2021_learning} or a state-based trained expert~\cite{song2023_learning}. Another line of work focuses on using the UAV or MAV for autonomous aerial cinematography or low altitude traffic surveillance that can reason about both geometry and scene context in real-time~\cite{bonatti2020autonomous, bozcan2020air}.

Unlike these works, we consider high-\dof systems (e.g., mobile robots with an onboard camera controlled by multiple joints) with kinematic constraints on environments with many obstacles, where optimization is likely to get stuck in a local minima. As such, we focus on planning simultaneously for efficient motion and perception quality on the high-dimensional space of robots such as mobile manipulators instead of UAVs or MAVs. Furthermore, unlike most of the perception-aware planning works that aim at maintaining visibility of landmarks and informative regions for visual odometry, we address more complex perception tasks which are significantly more computationally expensive to evaluate. 

\begin{table}[t]
\centering
\scriptsize
\setlength{\tabcolsep}{3.5pt}

\caption{
Comparison with active perception and perception-aware path planning. \textbf{Fixed Goal}: task specifies a fixed goal configuration. \textbf{Cont.\ Perc.}: perception constraints enforced along the whole trajectory, not just at waypoints. \textbf{General Perc.}: supports general perception objectives beyond a single task category. \textbf{High-DoF}: applicable to high-degree-of-freedom robots.
}
\label{tab:comparison}
{\color{black}
\begin{tabular}{@{}lcccc@{}}
\toprule
\textbf{Method} & \textbf{Fixed Goal} & \textbf{Cont. Perc.} & \textbf{General Perc.} & \textbf{High-DoF} \\
\midrule
Ours & \cmark & \cmark & \cmark & \cmark \\
Active Perception & \xmark & \xmark & \cmark & \cmark \\
Perception-aware Planning & \cmark & \cmark & \xmark & \xmark \\
\bottomrule
\end{tabular}
}
\end{table}

\subsection{Active Perception}

Also referred as view planning, active perception considers applications such as object~\cite{mendoza2020supervised} and scene~\cite{yi2024view} reconstruction, object recognition~\cite{johns2016pairwise}, pose estimation~\cite{zeng2020view}, and active mapping~\cite{liu2021active}, with single robot or multiple robots~\cite{sukkar2019multi}. In these tasks, perception is typically incorporated into the path planning process by selecting optimal viewpoints to maximize the performance of the specific problem~\cite{bajcsy1988active, aloimonos1988active, soatto2009actionable}. One of the goals of active perception is active localization, which seeks to compute control actions
and trajectories that minimize the pose estimation uncertainty. Most active localization works have been in the context of simultaneous localization and mapping (SLAM) or exploration \cite{liu2021active, morilla2024perceptual}.  Depending on the sensor used, they can be classified into range-based (e.g. RGB-D cameras)~\cite{feder1999adaptive, bourgault2002information, bachrach2012estimation} or vision-based~\cite{davison2002simultaneous, vidal2010action, mostegel2014active, sadat2014feature, morilla2024perceptual}.

Viewpoint planning or next best view planning (NBV) is a subset of active perception where the sensor pose or a sequence of poses is planned to maximize the information gain, i.e., minimize the entropy or uncertainty about the state of the environment or target objects, subject to constraints such as obstacle avoidance and movement cost. The challenge of next best view planning is proposing and selecting views from which a scene can be efficiently observed. The quality of an observation can be quantified by its accuracy (i.e., how closely the captured data resembles the actual scene) and completeness (i.e., what proportion of the scene has been observed). Observation accuracy primarily depends on the sensor capabilities but can be improved by considering the scene texture and geometry. The completeness of an observation is determined by the coverage obtained from captured views. The efficiency of a NBV approach is quantified by the time and travel distance required to obtain an observation. For detail oriented tasks, especially at the object level, such as active recognition, pose estimation~\cite{atanasov2014nonmyopic}, mapping or reconstruction~\cite{li2005information, kriegel2015efficient}, manipulators or mobile manipulators are typically used with attention-driven next best view planning. 

Recently, learning-based active perception~\cite{tang2024active,wang2025observe,jin2025activegs,chen2025splat,xiao2025vision} has become popular across robotics. These approaches primarily formulate perception as an information-gathering problem, finding viewpoints that reduce uncertainty in mapping, object state estimation, or task-relevant observations. For example,~\cite{tang2024active} formulates damage inspection as a POMDP and uses deep reinforcement learning to select next sensing poses that maximize crack-segmentation quality during underwater structural inspection. The authors in~\cite{wang2025observe} propose an asynchronous active vision–action framework that serially trains a next-best-view camera policy and a next-best-pose gripper policy for manipulation tasks under occlusion, but the perception policy is still defined in a low-dimensional camera-space rather than full robot configuration space. Several mapping-driven approaches~\cite{jin2025activegs,chen2025splat} focus on mobile robots or aerial platforms, where active perception guides viewpoint selection to improve scene reconstruction fidelity.~\cite{jin2025activegs} uses confidence-driven next-best-views to actively densify Gaussian-splatting maps, while~\cite{chen2025splat} develops a real-time navigation pipeline in the maps that couples safe corridor planning with fast vision-based localization. Broader surveys such as~\cite{xiao2025vision} further emphasize how vision-based learning in drones centers around improving perception for navigation, mapping, and control, typically in 3-\dof mobile systems.

Unlike learning-based active perception methods which choose perception-oriented intermediate goals or viewpoints, our planner operates with a fixed motion planning goal. In addition, active perception is largely decoupled from whole-body motion planning and focuses on viewpoint selection for mapping or manipulation in low-dimensional action spaces. This is different from our objective of considering perception constraints directly within high-\dof motion planning, which requires sensing quality, occlusion reasoning, and task performance to be jointly evaluated in the robot’s full configuration space. In~\cref{tab:comparison}, we summarize the different objectives of our method, active perception and perception-aware path planning.

\subsection{Implicit Neural Representations/Neural Surrogate Model}

A surrogate model is an engineering approach employed when directly measuring or computing an outcome of interest is either infeasible or prohibitively expensive. In such cases, an approximate model of that outcome is used instead~\cite{razavi2012review}. With the rise of deep learning, neural or deep learning-based surrogate models have been applied in numerous domains, including earth science~\cite{weber2020deep}, image processing~\cite{xiao2021conditioning}, fluid dynamics~\cite{du2022deep}, and power grids~\cite{hamid2023deep}. In robotics and perception, recent advances in machine learning have produced effective implicit neural representations of spatial information, such as Neural Radiance Fields (NeRFs)~\cite{mildenhall2021nerf} and Signed Distance Fields~\cite{park2019deepsdf}. These representations have been used for tasks ranging from learning multi-object dynamics~\cite{driess2023learning} and defining manipulation planning constraints~\cite{driess2022learning}, to enabling reactive robot manipulation~\cite{koptev2022neural} and vision-based navigation~\cite{adamkiewicz2022vision}. They offer compact, efficient storage~\cite{adamkiewicz2022vision}, a continuous geometric representation~\cite{driess2022learning, adamkiewicz2022vision, koptev2022neural}, and can be learned directly from sensor data~\cite{camps2022learning}, making them particularly appealing for planning. Moreover, recent years have seen neural representation models trained on broad datasets demonstrate impressive potential for generalizable manipulation~\cite{huang2023voxposer, simeonov2022neural, shridhar2022cliport, wen2022you}. For instance, neural object descriptors (NODs)~\cite{simeonov2022neural} have emerged as a powerful method for extracting dense, part-level features that generalize across object instances. Building on these directions, our work trains a neural model to map robot configurations directly to the desired perception metric (e.g., object detection confidence) as it is infeasible and very expensive to compute the perception score for different robot configurations in the real world.

\begin{figure}
    \centering
    \includegraphics[width=0.78\columnwidth]{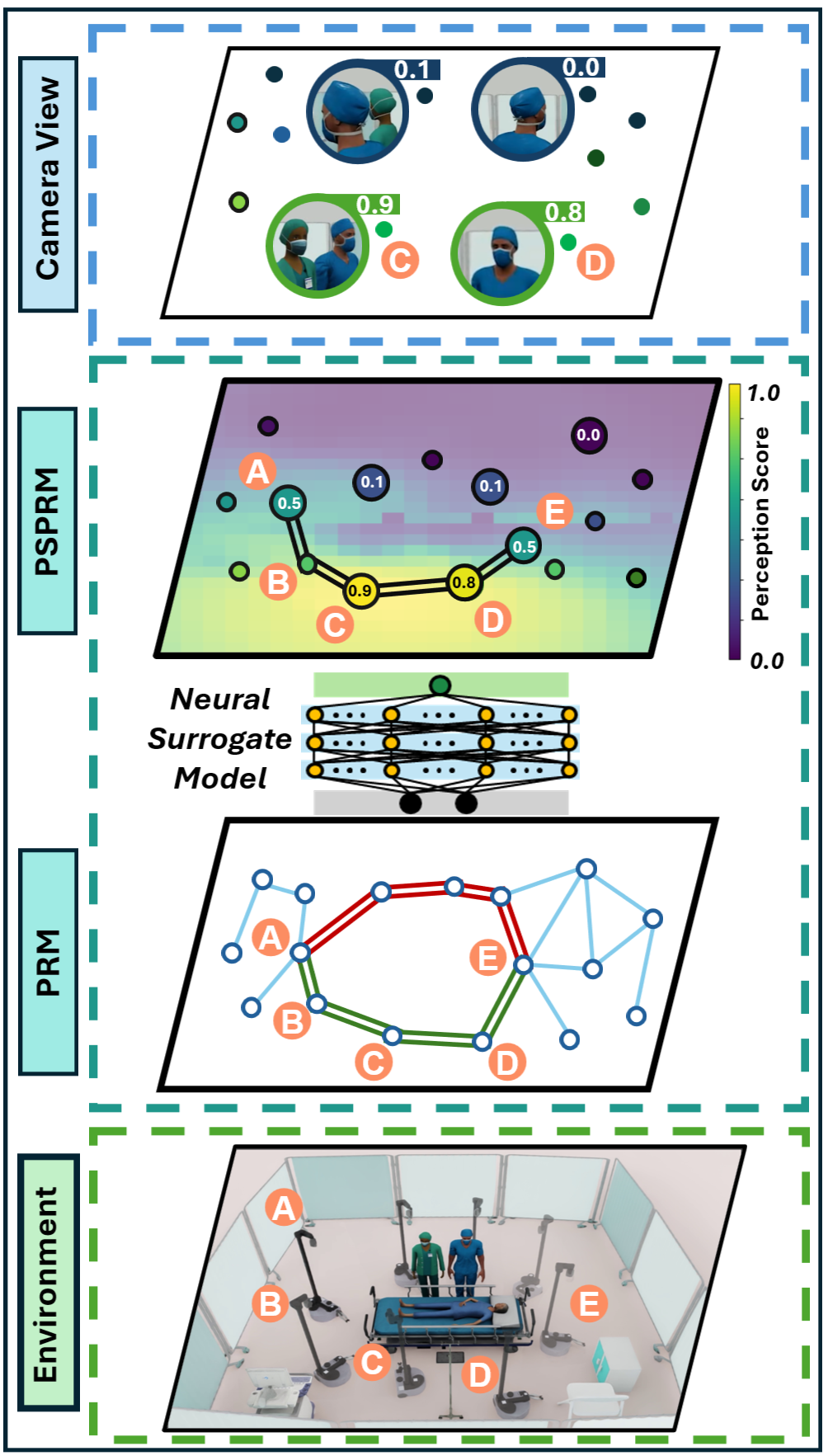}
    \caption{The PS-PRM pipeline. \textbf{Bottom:} environment with candidate configurations. \textbf{Middle:} PRM generation, neural surrogate scoring, and occlusion checking run in parallel to select the optimal path. \textbf{Top:} camera views confirming differing perception scores.
}
\label{fig:psprm_diagram}
\end{figure}

\section{Overview of PS-PRM}
\label{sec:method_overview}

We propose PS-PRM, a perception-constrained roadmap-based planning framework for high-\dof robots that jointly optimizes motion feasibility and task-dependent perception quality. PS-PRM is designed to operate in both static and dynamic environments, enabling robots to maintain high-quality perception of task-relevant objects or humans while performing navigation or manipulation tasks.

\cref{fig:psprm_diagram} illustrates the overall PS-PRM pipeline, including roadmap construction, perception score estimation, and path selection based on combined motion and perception costs. At a high level, PS-PRM constructs a PRM over a manifold defined by object-centered visibility constraints, such as line of sight or camera frustum alignment. Each configuration in the roadmap is scored based on its estimated perception quality, using a learned neural surrogate model. To handle cluttered scenes and inter-object occlusions, we further refine the perception score using a ray-casting-based occlusion checking module. Finally, PS-PRM selects the path that maximizes perception quality while minimizing motion cost. We choose PRM over tree-based planners in this work because it simplifies the underlying optimization problem and allows for more efficient node sampling and evaluation in batch, which can be efficiently parallelized on the GPU.

In dynamic environments, where objects or people may move, PS-PRM supports rapid replanning through a GPU-accelerated pipeline that updates perception scores, re-evaluates roadmap validity, and generates new plans in real time. The system is designed to operate efficiently at scale by batching key operations such as forward kinematics, surrogate model inference, and collision checking in environments where radical changes are not expected.

The following sections describe the three main components of PS-PRM in detail. 
\cref{sec:method_neural-surrogate-model} introduces the neural surrogate model used to estimate perception scores. \cref{sec:prm_planner} describes the construction of the perception-score-guided planner. \cref{sec:replanning} presents the GPU-parallelized pipeline.

\begin{figure}
    \centering
    \includegraphics[width=0.85\columnwidth]{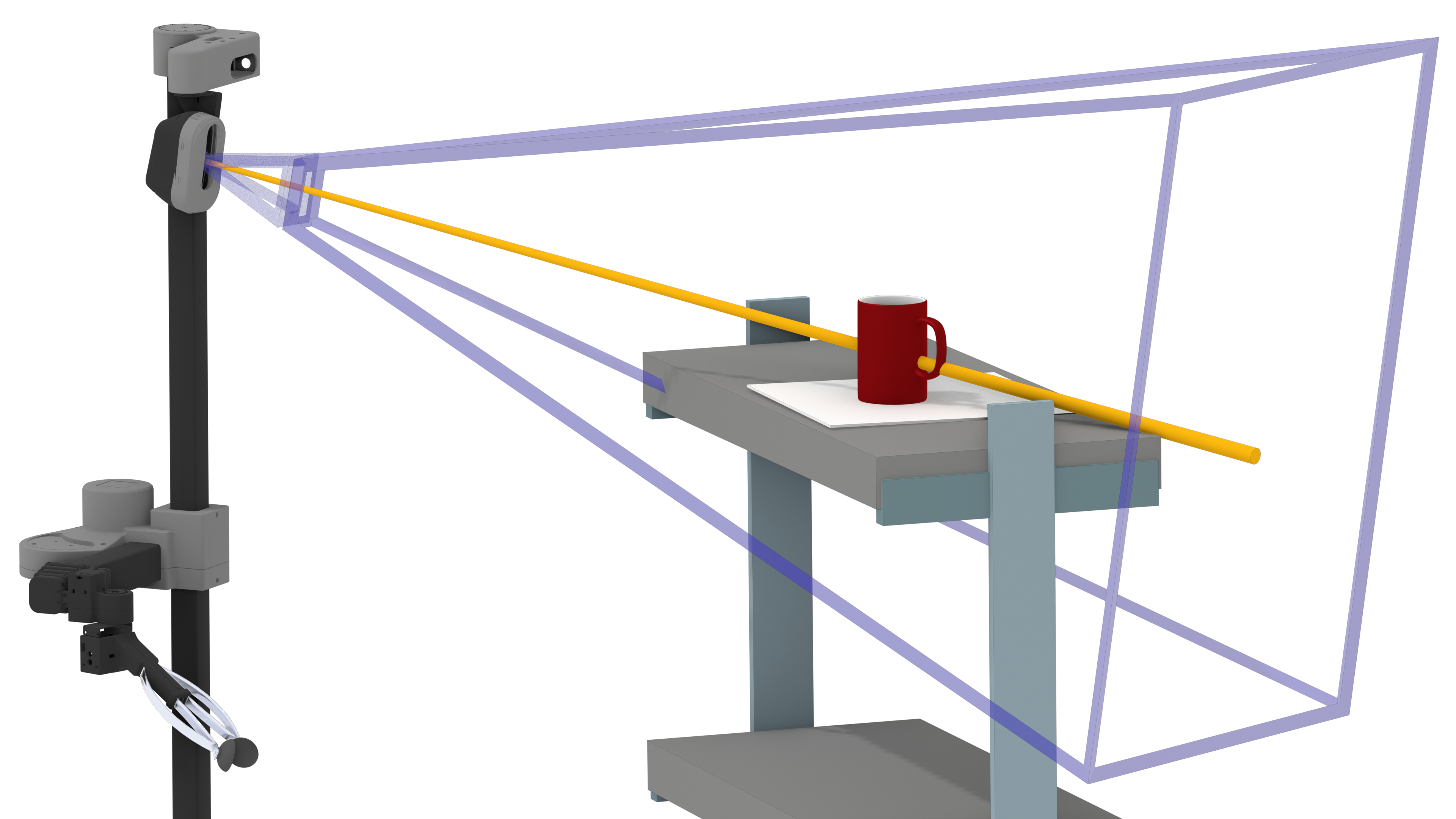}
    \caption{The camera model: line of sight (orange) and camera frustum (purple), shown against a cup intersecting each.
}
\label{fig:camera_model}
\end{figure}

\begin{figure*}[!t]
    \centering
    \includegraphics[width=0.9\textwidth]{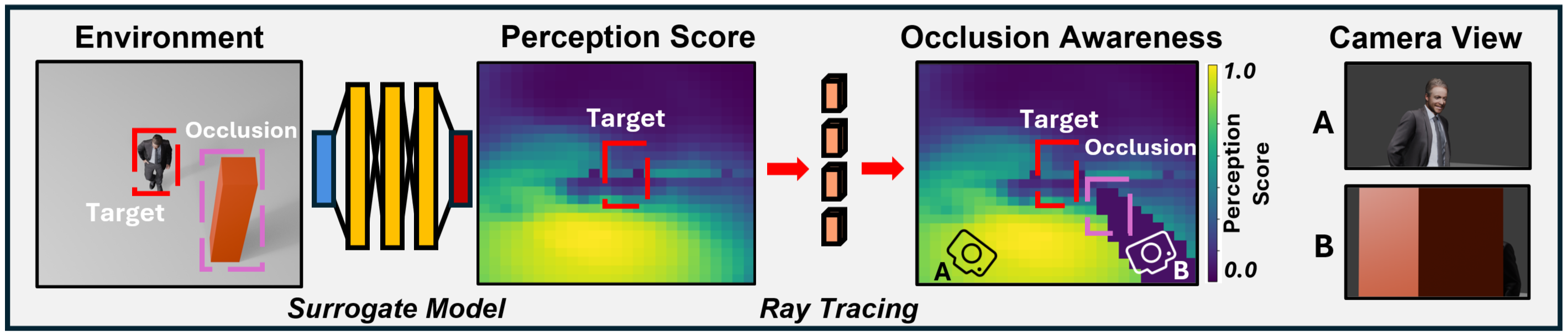}
    \caption{Perception-aware evaluation via neural surrogate model and occlusion-aware ray tracing. Left to right: environment with target (human) and occluder (orange box); surrogate-predicted perception score field; ray-traced occlusion refinement; camera views at points A and B, showing lower scores under occlusion.}
\label{fig:occlusion_heatmap}
\end{figure*}

\subsection{Neural Surrogate Perception Score Models}
\label{sec:method_neural-surrogate-model}

In real-world environments, it is infeasible to directly evaluate perception scores for all robot configurations due to the time and effort required to physically reposition the robot and collect sensor data at scale. As a result, simulation offers a practical alternative: scoring each configuration by rendering the robot’s camera view and passing it through a perception model. However, this approach is still computationally intensive. Each evaluation involves placing the robot in the simulated scene, generating a photorealistic rendering, and running the task-specific perception model to obtain a score. As demonstrated in our experiments, rendering high-quality images is time-consuming, especially when evaluating a large number of robot configurations. Moreover, perception scores are environment-dependent; any change in the object requires re-scoring previously scored configurations.

To address these challenges, we propose a neural surrogate model inspired by~\cite{quintero2024stochastic, long2025neural} that approximates perception scores without requiring explicit rendering. \cref{alg:neural_surrogate} outlines the full pipeline. Given a robot configuration and the target object label, the surrogate model predicts the expected perception performance (e.g., detection confidence for a cup). We first apply FK to map the configuration from \cspace\ to the robot’s camera pose. The neural model is then trained to map these camera transforms to perception scores obtained from labeled, photorealistic simulations. Specifically, we generate a dataset by rendering views of objects or humans from varied camera poses in simulation and evaluating each image with a perception model. The surrogate is implemented as a multilayer perceptron (MLP)~\cite{rumelhart1986learning} that takes the relative pose between camera and object, along with an object encoding, and outputs the corresponding perception score. Trained on this dataset, the surrogate enables fast and scalable evaluation during planning. Importantly, because the input is based on camera-to-object pose rather than robot-specific joint values, the model generalizes across robots and environments with similar perception setups. 
This robot-agnostic formulation is central to our approach: by operating in SE(3) camera space rather than robot configuration space, the surrogate can be reused across different platforms without retraining.

To address object inter-occlusion in cluttered indoor environments, we build upon techniques similar to those in \cite{xu2024makeway, gao2024relightable} by performing ray-casting towards surrounding objects. This step determines whether the object of interest is occluded by checking for intersections between each cast ray and the Axis-Aligned Bounding Boxes (AABB) of nearby objects. For a new robot configuration, we obtain an occlusion score in addition to the perception score estimated by the neural surrogate model. This occlusion score is computed as the ratio of rays that ultimately do not intersect the bounding box of the target object, an indicator that something else is blocking the view. If the occlusion score exceeds a chosen threshold, we set the perception score of the neural surrogate model to zero for that configuration. \cref{fig:occlusion_heatmap} shows a detailed overview of this occlusion checking step.

\begin{figure*}
    \centering
    \includegraphics[width=0.8\textwidth]{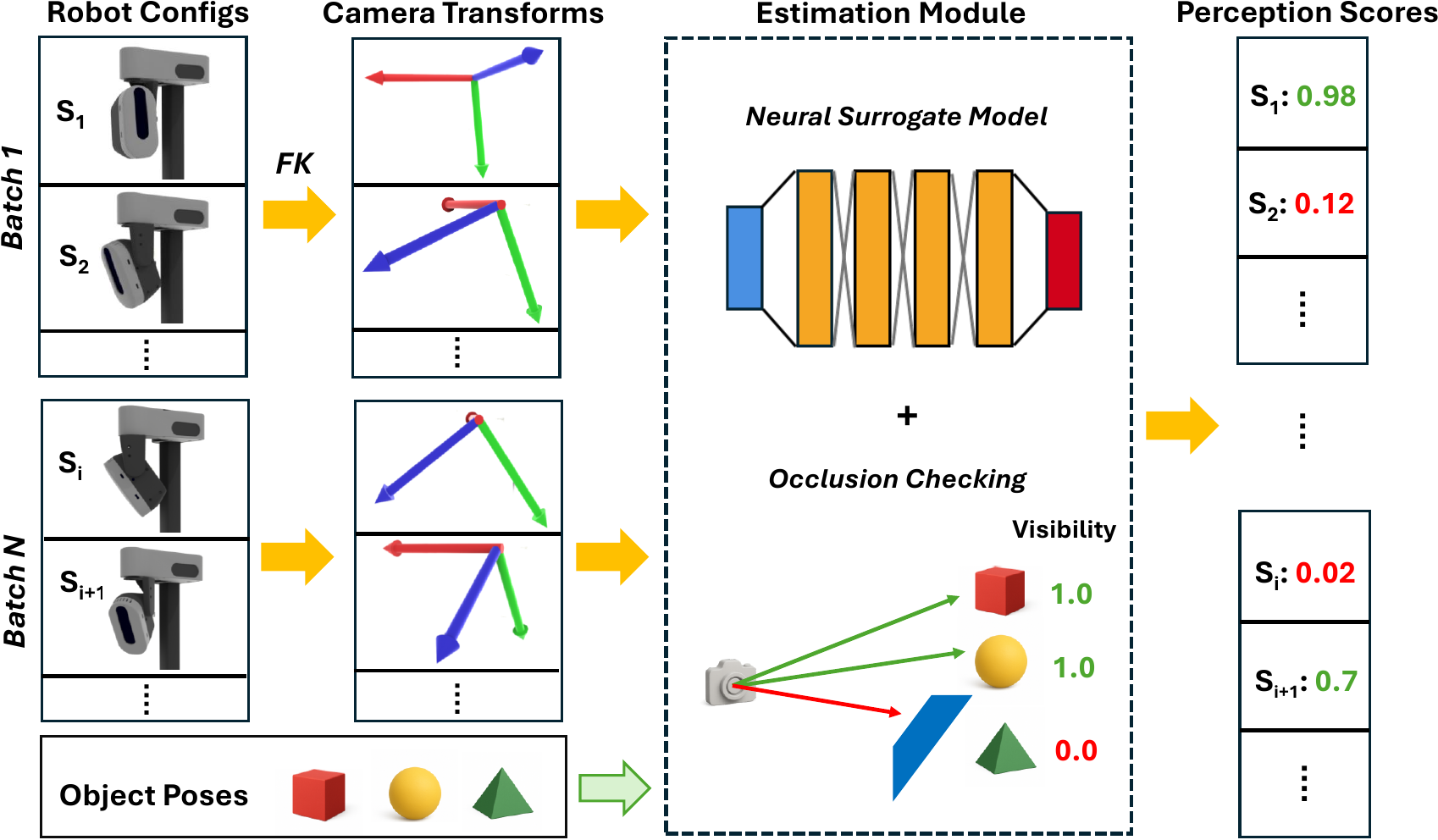}
    \caption{Parallel perception-score evaluation for a batch of robot configurations. Camera poses are computed via FK; occlusion checking and neural surrogate scoring run concurrently; outputs combine into a final perception score (0--1) per configuration.}
    \label{fig:neural_model_flowchart}
\end{figure*}

\cref{fig:neural_model_flowchart} illustrates the overall neural surrogate perception score model. This two-tier approach ensures that the neural surrogate model also accounts for occlusion in an efficient, fast manner, since ray-casting against bounding boxes can be highly parallelized. In practice, relying on bounding boxes provides a reasonable approximation of object geometry, though it may not perfectly capture complex shapes or partial occlusions. Nevertheless, for many motion-planning tasks in cluttered environments, it strikes a good balance between speed and accuracy.

\subsection{Perception Score-Guided Planner}
\label{sec:prm_planner}
To enable robots to maintain a high score for a given perception task throughout their motion, the PS-PRM planning process consists of two key stages: (1) sampling configurations that satisfy visibility constraints, and (2) constructing, scoring, and planning over a PRM built from these configurations. The first stage ensures that each sampled configuration allows the robot to keep the target object or person within view. The second stage integrates perception quality scores into the roadmap and performs graph-based planning to find a path that balances motion efficiency and perception quality. We describe each stage in detail below.

\subsubsection{Sampling with Visibility Constraints}
\label{sec:method_baseline}
To maximize perception scores along the planned motion, the robot’s onboard camera must continuously keep the object or person of interest in sight. In this work, we consider two commonly used visibility constraints for maintaining such monitoring, as illustrated in \cref{fig:camera_model}: line of sight and camera frustum. In the line-of-sight model, the robot’s optical axis is aligned with the centroid of the target, effectively keeping the target centered on the image plane (\eg using a pinhole camera model)~\cite{falanga2018_pampc, tordesillas2022_panther}. In the camera frustum model, the objective is to keep the target within the camera’s viewing volume, represented as a rectangular pyramid or frustum~\cite{sung2008essentials}. These visibility constraints define a manifold within the robot’s free configuration space consisting of configurations where the target remains in view. 

More formally, we define the set of all robot configurations that enable the robot to monitor the object or person of interest to lie in a manifold \(M \) implicitly represented as the zero-level set defined by \( h_{M}(q) = 0 \) where $q \in \cspace$, \( h_{M}(q) : \mathbb{R}^k \rightarrow \mathbb{R}^l \), and \( l \) is the intrinsic dimensionality of the constraint (\( l \leq k \)). For the line-of-sight model, $h_{M}(q)$ is the distance between the centroid of the object and the line of sight from the robot's camera. For the camera frustum model, we model $h_{M}(q)$ as $\max(0,~\text{sd}_f(q, p))$ where $\text{sd}_f$ is the signed distance between the object $p$ and the surface of the robot's camera frustum when the robot is at configuration $q$. 

For high-\dof robots, computing motion that considers the aforementioned object-centered models can be achieved using sampling-based motion planning (SBMP)~\cite{orthey2023sampling} with manifold constraints~\cite{kingston2018sampling} where planning happens on a subspace of the free configuration space defined by a manifold. Specifically, in the construction of our proposed planner shown in Alg. \ref{alg:psprm}, \psprm first constructs a PRM with states and transitions that satisfy the manifold constraints, using either line-of-sight or camera frustum. The PRM is defined $G=(V,E)$, where $V$ is the set of configurations and $E$ the set of edges connecting them. The neighbor nodes in the PRM are constrained to lie on the same manifold and are selected based on their geodesic proximity within a certain distance. To sample from the manifold, \psprm first selects a robot configuration from \cfree and then applies the projection operator to map it onto the surface of \(M \). This involves iteratively retracting the point towards a minimum of the constraint function and solving a linear system of equations at each step. Configurations that cannot be projected into valid in-manifold configurations are discarded, and the algorithm resamples until a valid configuration is found or a timeout is reached. This process ensures that the object is placed either within the line of sight or inside the camera frustum.

While enforcing visibility constraints helps keep the target object or person within the robot’s sight, this alone is not sufficient for many real-world perception tasks. Our experiments show that visibility-based approximations often fall short in practice as they lack semantic and geometrical information specific to the objects and the task. This affects in complex environments where factors such as object shape (e.g., human faces looking clearer closer and in front of the person), occlusions, background clutter, and lighting conditions significantly affect perception performance~\cite{yin2021center, zheng2023deep, minaee2021image}. To address these limitations, our proposed planner incorporates task-aware perception scores to guide planning toward configurations that support more reliable perception.

\subsubsection{PRM construction and scoring}
\label{sec:method_prm}
While the visibility constraints provide a bound on necessary conditions for the perception task, they do not consider the actual performance of the task. To remedy this, \psprm expands the Probabilistic Roadmap planner (PRM)~\cite{kavraki1996probabilistic}, by incorporating perception scores into the graph and using them during planning. 

After \psprm builds the manifold-constrained PRM, the edges initially encode only the distance between configuration nodes under a given metric (e.g., Euclidean distance from state to state). This is used as the motion cost, normalized to facilitate its later combination with the perception cost. \psprm then assigns perception scores to the edges to account for the perception task. 
Since each perception score is calculated discretely for each state, thus ignoring the continuous path taken from node to node, \psprm also uses interpolated intermediate configurations to compute a normalized perception score for an edge, $c_p(q_u, q_v) \in [0,1]$. \psprm performs 
robot-space dependent interpolation between the $q_u$ and $q_v$ to generate $N$ valid intermediate configurations, $q_1, q_2, \ldots, q_N$. It then 
calculates the perception scores corresponding to these intermediate configurations
and computes their normalized perception score $c_p(q_1), c_p(q_2), \ldots, c_p(q_N)$. Next, \psprm calculates the average normalized perception score of the two configurations connected by an edge and their intermediate configurations to obtain the normalized perception score associated with the edge $c_p(q_u, q_v)$:
\begin{equation}
\label{normalized_pscore}
c_p(q_u, q_v) = \frac{c_p(q_u) + c_p(q_v) + \sum_{i=1}^{N} c_p(q_i)}{N+2}
\end{equation}

Finally, \psprm linearly combines the normalized perception score and normalized motion cost for an edge to determine a new edge weight. While other nonlinear combinations are explored, the linear formulation offered a good balance between interpretability and empirical performance. Since the perception score is intended to be maximized while the motion cost is intended to be minimized, we subtract the normalized perception score and add a constant $1$ to ensure that the edge weight remains non-negative, within the range $[0, 2]$. The combination is defined as follows, where $\alpha \in [0, 1]$ is a factor that balances the perception score and motion cost:
\begin{equation}
\label{method:combined_edge_weight}
c(q_u, q_v) = 1 - \alpha \cdot c_p(q_u, q_v) + (1 - \alpha) \cdot c_m(q_u, q_v)
\end{equation}

After the PRM is constructed, \psprm connects the start and goal to the roadmap and applies an A* search~\cite{hart1968formal} with an admissible heuristic to find a path from $\cstart$ to $\cgoal$ for the robot to follow.

Next, to guarantee the optimality of the A* search, we show that our A* heuristic remains admissible under the modified edge-cost formulation. Since consistency implies admissibility~\cite{choset2005principles}, it is sufficient to prove that our heuristic is consistent. We define our A* heuristic for a PRM $G=(V,E)$ as a ``hop"-based function that lower-bounds the remaining motion and perception costs:
\begin{equation}
h(q) \;=\; H_{\min} \cdot c^{\min},
\label{eq:heuristic_def}
\end{equation}
where $H_{\min}(q)$ denotes the hop distance, i.e., the minimum number of edges required to reach the goal from the configuration $q$  obtained via a shortest-path search, and $c^{\min}$ represents the minimum edge cost over the entire roadmap:
\begin{equation}\label{eq:min_edge_def}
c^{\min} = \min_{(u,v)\in E} c(q_u, q_v).
\end{equation} 

For any edge~$(u,v)\in E$,~a~feasible~path~from $u$ to the goal is to first take $(u,v)$ and then follow a shortest path from $v$ to the goal. Thus, we have: 
\begin{equation}
H_{\min}(q_u) \;\le\; 1 + H_{\min}(q_v).
\label{eq:hop_ineq}
\end{equation}
Therefore, our heuristic function $h(\cdot)$ satisfies:
\begin{align}
h(q_u)&= H_{\min}(q_u) \cdot c^{\min}
        &&\text{\cref{eq:heuristic_def}} \nonumber\\
     &\le (1+H_{\min}(q_v))\cdot c^{\min}  
        &&\text{\cref{eq:hop_ineq}} \nonumber\\
     &\le c(q_u,q_v) + h(q_v)
        &&\text{\cref{eq:heuristic_def,eq:min_edge_def}} \nonumber
\end{align}
 which is exactly the condition of a consistent heuristic. Thus, we show that our A* heuristic remains consistent and admissible under the modified edge cost formulation.

This formulation only requires nonnegativity of edge costs, without assuming a specific form such as Euclidean distance for motion cost. It applies broadly, e.g., when $c_m$ is defined as the trajectory length, energy, or control effort, and $c_p$ is a non-negative normalized neural perception score. In this paper, we demonstrate experiments on object detection and tracking, as well as human face detection and activity recognition, though the approach can be generalized to other perception tasks.

Since our problem is formulated as a sampling-based motion planning problem, probabilistic completeness~\cite{choset2005principles} is an important property. We therefore show that probabilistic completeness still holds after introducing our modified edge-cost formulation. Specifically, we show that our perception-weighted PRM formulation does not invalidate the standard PRM completeness. This holds because the reweighting affects only the edge costs, not the set of feasible configurations or connections. Assume a PRM constructs a graph $G = (V,E)$ over samples from the collision-free configuration space $Q_{\mathrm{free}}$. The PRM completeness proof on Theorem 7.4.1 from~\cite{choset2005principles}, does not depend on edge cost function, only on samples and connections. Let $\sigma$ denote any feasible path with clearance $\delta > 0$. With a probability exponentially increasing with the number of sampled states, PRM will include vertices $\{q_1,\ldots,q_n\}$ such that consecutive vertices are within a local planner’s connection radius and $\sigma = (q_1,\ldots,q_n)$ will be found. Since our composite edge cost $c(u, v)$ shown in \cref{method:combined_edge_weight}
is nonnegative and finite for all $(u,v)\in E$, the reweighting does not eliminate any edges or feasible connections.  Consequently, PS-PRM remains probabilistically complete: if a feasible path exists in $Q_{\mathrm{free}}$, the probability that PS-PRM finds a finite-cost path approaches one as the number of samples increases.

\begin{algorithm}
\caption{Perception Score-Guided PRM Construction (\texttt{PS-PRM})}
\label{alg:psprm}
\begin{algorithmic}[1]
\State \textbf{Input:} Environment $\mathcal{E}$, Robot Model $\mathcal{R}$, Object of Interest $\mathcal{O}$
\State \textbf{Output:} Roadmap $\mathcal{G} = (V, E)$ with perception-aware edge weights
\State Initialize empty graph $\mathcal{G} = (V, E)$
\State $N \gets$ number of samples
\State \textbf{for} $i = 1$ to $N$ \textbf{do}
    \State \quad Sample $q_i$ in free space $\mathcal{C}_{\text{free}}$
    \State \quad Project $q_i$ onto the manifold using constraint projection
    \State \quad \textbf{if} $h_M(q_i)$ is less than a threshold \textbf{then}
        \State \quad \quad Add $q_i$ to $V$
    \State \quad \textbf{end if}
\State \textbf{end for}
\State \textbf{for} each $q_u \in V$ \textbf{do}
    \State \quad \textbf{for} each $q_v$ within geodesic distance $\delta$ \textbf{do}
        \State \quad \quad Normalize the original distance-based edge weight between $q_u$ and $q_v$ to obtain the motion cost $c_m(q_u, q_v)$.
        \State \quad \quad Generate $N$ intermediate configurations $q_1, \dots, q_N$
        \State \quad \quad Compute normalized perception score $c_p(q_u, q_v)$ using Eq.~(\ref{normalized_pscore})
        \State \quad \quad Compute edge weight $c(q_u, q_v)$ using Eq.~(\ref{method:combined_edge_weight})
        \State \quad \quad Add edge $(q_u, q_v)$ to $E$ with weight $c(q_u, q_v)$
    \State \quad \textbf{end for}
\State \textbf{end for}
\State \Return $\mathcal{G}$
\end{algorithmic}
\end{algorithm}

\begin{algorithm}
\caption{Neural Surrogate Model for Perception Score Estimation}
\label{alg:neural_surrogate}
\begin{algorithmic}[1]
\State \textbf{Input:} Robot Configuration $q$, Object $\mathcal{O}$
\State \textbf{Output:} Perception Score $c_p(q)$
\State Compute camera transform $T_c$ using forward kinematics
\State Query neural surrogate model: $c_p(q) \gets \text{NN}(T_c, \mathcal{O})$
\State Perform ray-casting to check occlusions
\State \textbf{if} occlusion score $> \tau$ \textbf{then}
    \State \quad Set $c_p(q) \gets 0$
\State \textbf{end if}
\State \Return $c_p(q)$
\end{algorithmic}
\end{algorithm}

\subsection{GPU-Parallelized Replanning Pipeline}
\label{sec:replanning}

The proposed perception scoring methods are highly parallelizable. Since the perception scores are computed using a neural surrogate model specifically optimized for GPU inference, we are motivated to implement the entire planning pipeline on the GPU. For collision checking, PS-PRM evaluates the validity of sampled configurations with hardware acceleration: instead of checking each obstacle individually, the algorithm performs batched collision checking on all obstacles simultaneously. Furthermore, both FK and surrogate model inference are performed in batches and executed in the GPU. This node-level parallelization accelerates roadmap construction and perception-scoring and enables real-time replanning in high-dimensional spaces. PS-PRM can therefore operate effectively in dynamic environments where the object of interest, goals or obstacles may move during execution. In such cases, the planner updates node perception scores in real time and replans the path as needed to adapt to the new environment. \cref{fig:neural_model_flowchart} shows a detailed overview of the GPU-accelerated process.

As the robot executes its planned trajectory, it continuously monitors whether the object has moved. If movement is detected (\eg by the onboard camera), PS-PRM invokes the parallelized pipeline to re-project the existing nodes and update their perception scores.

Nodes that cannot be projected into valid configurations or that collide in the new environment are pruned from the roadmap along with their associated edges. After pruning and re-scoring, the planner re-evaluates the current trajectory by computing a new average perception score; if that score falls below a specified threshold, it triggers a replanning step to find a more suitable path from the robot’s current position to the goal region. If the remaining nodes fail to produce a safe path to the goal, PS-PRM partially rebuilds the roadmap by sampling additional nodes while retaining any valid ones, thus avoiding a complete reconstruction. This leverages each step based on their computational needs, while building a PRM can be performed in seconds, re-evaluating nodes' collisions and scores can be done at higher frequencies of several times a second. Through this GPU-accelerated parallelization, PS-PRM rapidly adapts to dynamic changes in the environment and maintains both collision-free and perception-optimized trajectories with reduced computation time.

In this work, we primarily consider small to medium target or obstacle motions, where changes in the environment are gradual relative to the planner’s replanning frequency. Under these conditions, the proposed planner can update trajectories online by replanning fast enough to adapt to dynamic targets, obstacles, and goals. Specifically, \cref{experiment:dynamic_environment} demonstrates the feasibility of our system. Large or abrupt target motions are not the primary focus of this work. 
When the target dynamics exceed the planner’s replanning capability (e.g., rapid, discontinuous, or adversarial movements), the problem becomes fundamentally one of high-frequency tracking or reactive control rather than perception-constrained motion planning, which is outside the scope of this paper.
\section{Experimental Results}
\label{sec:experiment}
In this section, we evaluate our proposed methods in simulation and real-robot experiments using the Hello Robot's Stretch $2$~\cite{kemp2022design}, Fetch robot, and a UR5. We use PyBullet~\cite{coumans2021} for collision checking, \ompl's Python
bindings~\cite{sucan2012open} for planning, and Isaac Sim~\cite{nvidia2022isaacsim} for simulation and visualization.
All experiments were conducted on an Intel i7-12700K CPU and an NVIDIA GeForce RTX4090 GPU. We consider both object detection/tracking, human face detection and activity recognition as perception tasks. For the robot's camera parameters in simulation experiments, we use a resolution of 640 $\times$ 480 pixels, a field of view of 42.5 degrees, and a clipping range of 0.3 meters for the near plane and 10 meters for the far plane.

\begin{table*}
\centering
\caption[Performance Metrics]{Planning and perception results for the Stretch robot in a simulated home environment, averaged over 2,500 planning problems. ``Preprocess/Plan./Total Time'' are pre-planning, planning, and combined durations; ``Path Len.'' and ``Succ. Rate'' are motion planning metrics; ``Det. Rate,'' ``Conf. Score,'' and ``Track Rate'' are object-tracking metrics.}
\label{tab:planning-results}
\resizebox{\textwidth}{!}{
\begin{tabular}{@{}c|ccccc|ccc@{}}
 & \multicolumn{5}{c}{Motion Planning}  & \multicolumn{3}{c}{Object Tracking} \\
Method & Preprocess Time (s) & Plan. Time (s) & Total Time (s) & Path Len. (rad) & Succ. Rate & Det. Rate  & Conf. Score  & Track Rate  \\ \midrule
\rowcolor{gray!10} Post & - & 0.051 & 0.051 & 6.277 & 99.7\% & 20.9\% & 0.531 & 57.9\% \\
\rowcolor{gray!25} Rejection & - & 1.264 &  1.264& 10.379 & 98.9\% & 29.7\% & 0.599 & 74.0\% \\
\rowcolor{gray!35} Manifold-frustum & - & 0.495 & 0.495 & 11.141 & 98.8\% & 41.0\% & 0.597 & 80.9\% \\
\rowcolor{gray!35} Manifold-line & - & 2.859 & 2.859 & 7.478 & 90.1\% & 48.9\% & 0.482 & 67.4\% \\
\rowcolor{gray!50} Visibility-Aware TrajOpt & - & 0.152 & 0.152 & 5.62 & 93.0\% & 35.7\% & 0.561 & 64.4\% \\
\rowcolor{gray!50} RL-Based Active Perception & 169.8 & 0.404 & 170.204 & 21.75 & 57.3\% & 75.5\% & 0.634 & 80.0\% \\
\rowcolor{gray!70} \psprm-frustum & 291.7 & 0.014 & 291.714 & 16.688 & 96.4\% & 73.7\% & 0.761 & 89.8\% \\
\rowcolor{gray!70} \psprm-line & 341.5 & 0.018 & 341.518 & 15.007 & 95.7\% & \textbf{80.1\%} & \textbf{0.792} & \textbf{91.0\%} \\
\rowcolor{gray!70} Neural-\psprm-frustum & 32.08 & 0.017 & 32.097 & 15.814 & 97.1\% & 72.8\% & 0.759 & 89.8\% \\
\rowcolor{gray!70} Neural-\psprm-line & 31.79 & 0.021 & 31.811 & 13.518 & 96.3\% & 79.6\% & 0.771 & 90.4\% \\

\end{tabular}
}
\end{table*}

\subsection{Evaluated Methods}
\label{experiment:evaluated_ethods}
We implement four baseline methods as follows:
\begin{itemize}
    \item \textbf{Baseline 1 Post:} This baseline method initially performs standard motion planning without accounting for any constraints, and then post-processes the robot's free joints (\emph{e.g.,}~camera pan and tilt) along the resulting path to align the robot's line of sight with the centroid of the object or person as described in details in \cref{sec:method_baseline}. This is achieved through a simple geometric analysis given the position and orientation of the robot at each configuration with respect to the monitoring object or person.
    \item \textbf{Baseline 2 Rejection:} 
    This baseline method extends a standard sampling-based planner by modifying its feasibility function to ensure that sampled configurations remain within the robot's camera frustum.
    \item \textbf{Baseline 3 Manifold-line:} This baseline method is an implementation of a manifold constrained motion planner using the distance between the centroid of the object and the line of sight from the robot's camera's as $h_{M}(q)$ as described in details in \cref{sec:method_baseline}.    
    \item \textbf{Baseline 4 Manifold-frustum:} This baseline method is an implementation of the manifold constrained motion planner described in \cref{sec:method_baseline} using $h_{M}(q) = \max(0, \text{sd}_f(q, p))$ where $\text{sd}_f$ is the signed distance between the object $p$ and the surface of the robot's camera frustum when the robot is at configuration $q$. 
    \item \textbf{Baseline 5 Visibility-Aware TrajOpt:} This method draws elements from related trajectory optimization work \cite{wang2021visibilitytrajopt}, modeling the problem as a minimization of costs to reach the goal configuration, generate a smooth and kinematically feasible path and keep the object within view of the camera. Additional constraints are used to generate a non-holonomic trajectory for the mobile base and to ensure collision-free paths.
     \item \textbf{Baseline 6 RL-Based Goal-Conditioned Active Perception:}
    This method formulates perception-aware navigation problem as a goal-conditioned
    active perception problem and learns a policy via Proximal Policy
    Optimization (PPO)~\cite{schulman2017ppo}. This baseline is implemented based on elements from the works~\cite{jayaraman2018lookaround,mirowski2018learningtonavigate, tang2024active,wang2025observe}, where an agent jointly optimizes mobility and observability of a target object, and by standard goal-conditioned navigation~\cite{zhu2017targetdriven}.
\end{itemize}

The first four baseline methods are tree-based planners derived from RRT-Connect, and therefore they do not require preprocessing time.

The fifth baseline is a Visibility-Aware Trajectory Optimization planner, based on a drone-oriented related world~\cite{wang2021visibilitytrajopt}. Trajectories are modeled as a set of discrete points over which costs are calculated, accounting for joint position and velocity constraints, collision avoidance and angle of vision via forward kinematics. The optimization is modeled as a least squares problem and solved through Levenberg-Marquardt.

For the RL-Based Goal-Conditioned Active Percpetion baseline, the agent observes the current robot configuration, goal configuration, monitored object position, and episode progress, and outputs continuous configuration-space actions. Invalid actions are rejected using OMPL collision checking and penalized during training. The reward combines perception quality, motion cost, goal progress, collision penalties, and sparse goal completion rewards, encouraging the agent to jointly optimize navigation efficiency and target observability. Perception quality is approximated using a geometry-based proxy derived from forward kinematics, adapted from the objective function used in the ~\cite{tang2024active}. Similar to the works~\cite{tang2024active,wang2025observe}, the policy is parameterized by a shared actor-critic MLP with layers $[256,256,128]$ and trained using PPO with learning rate $3\times10^{-4}$, rollout length $512$, batch size $64$, and $10^5$ environment interaction steps. Training is performed on randomized start and goal configurations sampled from the environment, and the learned policy is evaluated deterministically for up to $200$ steps per query.

We implement four versions of our approach \psprm, as described in~\cref{sec:method_overview}, varying along two key dimensions: the perception model (line-of-sight vs. camera frustum) and whether a neural surrogate is used for scoring.

\begin{itemize}
\item \textbf{PS-PRM-line:} Uses a line-of-sight model to estimate perception quality, with direct evaluation (no neural surrogate).
\item \textbf{PS-PRM-frustum:} Uses a camera frustum model to estimate perception quality, with direct evaluation (no neural surrogate).
\item \textbf{Neural-PS-PRM-line:} Uses a line-of-sight model, with perception quality predicted by a neural surrogate model.
\item \textbf{Neural-PS-PRM-frustum:} Uses a camera frustum model, with perception quality predicted by a neural surrogate model.
\end{itemize}

For both \textbf{PS-PRM-line} and \textbf{PS-PRM-frustum}, we select $N = 5$ and $\alpha = 0.75$ in our experiments to strike a balance between action cost and perception performance. It is important to note that our planner is not highly sensitive to these parameters as long as they remain within a reasonable range. Through experimentation, we found that small variations in $N$ and $\alpha$ do not significantly impact the overall performance. 

For both \textbf{Neural-PS-PRM-line} and \textbf{Neural-PS-PRM-frustum}, we set the number of samples to $N = 5$ and the weight parameter to $\alpha = 0.8$ in our experiments, aiming to balance the trade-off between action cost and perception performance. To support this framework, we train a neural surrogate model that predicts perception scores for the relevant tasks.

\begin{table}[t]
\centering
\caption{In-distribution benchmark and out-of-distribution robustness of the neural surrogate model: estimation error (RMSE, MAE), ranking (Spearman's $\rho$), and detection classification (Accuracy, Precision, Recall, F1), under standard evaluation plus two OOD conditions (occlusion, dim lighting).}
\label{tab:benchmark-results}
\resizebox{\columnwidth}{!}{
\begin{tabular}{l cc | c | cccc}
\toprule
 & \multicolumn{2}{c}{Estimation Error} & \multicolumn{1}{c}{Ranking} & \multicolumn{4}{c}{Detection Classification} \\
\cmidrule(lr){2-3} \cmidrule(lr){4-4} \cmidrule(lr){5-8}
Method & RMSE $\downarrow$ & MAE $\downarrow$ & Spearman $\rho$ $\uparrow$ & Acc. $\uparrow$ & Prec. $\uparrow$ & Rec. $\uparrow$ & F1 $\uparrow$ \\
\midrule
\rowcolor{gray!10} \textbf{Surrogate Model} & 0.154 & 0.085 & 0.641 & 93.1\% & 74.4\% & 65.5\% & 69.7\% \\
\midrule
\multicolumn{8}{l}{\textit{Out-of-Distribution Robustness Analysis}} \\
\midrule
Occluded (11.8\% of valid configs)   & 0.258 & 0.151 & {---} & {---} & {---} & {---} & {---} \\
\midrule
Dim Lighting ($L/L_\text{ref} = 0.75$) & 0.049 & 0.034 & 0.641 & {---} & {---} & {---} & {---} \\
Dim Lighting ($L/L_\text{ref} = 0.50$) & 0.105 & 0.073 & 0.641 & {---} & {---} & {---} & {---} \\
Dim Lighting ($L/L_\text{ref} = 0.25$) & 0.175 & 0.122 & 0.641 & {---} & {---} & {---} & {---} \\
\bottomrule
\end{tabular}
}
\end{table}

Specifically, in our experiments, we use the object detection confidence score as the perception score for object detection and tracking tasks, as described in Sec.~\ref{experiment:benchmark_results}, Sec.~\ref{experiment:ur5_demo}, and Sec.~\ref{experiment:real_robot}. For human tracking tasks, such as those discussed in Sec.~\ref{experiment:nuring_experiment} and Sec.~\ref{experiment:dynamic_environment}, we use the confidence score of the human face detector RetinaFace~\cite{deng2019retinaface} from Deepface~\cite{serengil2024lightface} as the perception score.

The neural surrogate model is implemented as a Multilayer Perceptron (MLP)~\cite{rumelhart1986learning}, consisting of a shared architecture with five fully connected layers, each with 256 hidden units and ReLU (Rectified Linear Unit) activation. The input to the network includes the relative pose (comprising both position and orientation) between the camera and the object, as well as an object encoding that indicates which object the robot is monitoring. We use a one-hot encoding to represent the object being monitored. The network outputs the predicted perception score corresponding to that camera transform and object pose. 

This model is trained on a dataset composed of a diverse set of commonly encountered objects and humans from the COCO dataset~\cite{lin2014microsoft}, enabling generalization across a wide range of realistic detection scenarios. We generate a dataset of perception scores as the objects or human are viewed from different camera transform in Issac Sim. Specifically, we sample a set of 50000 camera transforms where the object or human are in the field of view uniformly at random. For each camera transform, we render the images and evaluate the perceptions scores using a perception model with the relevant perception tasks. Then, we train our model on the collected dataset for 200 epochs
with an Adam~\cite{kingma2014adam} optimizer, learning rate of $1 \times 10^{-4}$, and batch size of 512. 

For the dynamic experiments in \cref{experiment:dynamic_environment}, all evaluations are conducted using the GPU implementation, as the CPU-based version is not able to run and replan at a rate for dynamic scenarios. This highlights the necessity of fast replanning when targets, obstacles, or goals are moving, and demonstrates the practical advantage of GPU parallelism in enabling real-time perception-constrained planning.

\subsection{Perception Scoring Benchmark}

We evaluate the neural surrogate model from three complementary perspectives: confidence estimation accuracy, ranking consistency, and binary detection performance. Estimation quality is quantified using Root Mean Squared Error (RMSE) and Mean Absolute Error (MAE) between the predicted confidence scores and the YOLO-based ground truth, capturing absolute prediction error. Since the planner primarily depends on relative perception quality rather than exact confidence values, we additionally evaluate ranking consistency using Spearman's rank correlation coefficient~$\rho$, which measures how well the model preserves the ordering of camera poses according to perceptual quality. Finally, we assess binary perception score prediction using standard classification metrics (accuracy, precision, recall, and F1 score) with a confidence threshold of $0.75$.

As shown in Table~\ref{tab:benchmark-results}, the surrogate model achieves low estimation error, indicating accurate confidence prediction, while also maintaining a strong positive rank correlation, demonstrating its ability to preserve perceptual ordering across camera viewpoints. In addition, the high classification accuracy and balanced precision--recall performance indicate that the model reliably distinguishes high-quality perception configurations from lower-quality ones. Collectively, these results suggest that the surrogate provides both accurate and decision-relevant perception estimates, making it well suited for perception-guided motion planning.

To further evaluate generalization, we examine the surrogate model under two out-of-distribution conditions. For occlusion, we sample 500 valid robot configurations in the hospital scene and verify line-of-sight visibility using PyBullet raycasting. Among these samples, 11.8$\%$ of configurations contain object occlusion caused by surrounding furniture. Because the surrogate model operates solely on camera--object relative pose and does not explicitly encode geometric visibility, it systematically overestimates the perception score for occluded configurations, with a mean error of $0.151$ and an RMSE of $0.258$. Nevertheless, the planner's collision-free constraint implicitly mitigates part of this issue by excluding configurations in which the robot itself intersects with obstacles. As a result, the remaining 88.2$\%$ of sampled configurations maintain a good performance to the object.

For varying illuminance, the surrogate model is structurally invariant to lighting changes because its inputs are purely geometric. To characterize the expected degradation under low-light conditions, we model the confidence variation analytically using the power-law relationship $\hat{c}(L) = c_{\mathrm{nom}}(L/L_{\mathrm{ref}})^{0.6}$~\cite{michaelis2019benchmarking}, while the surrogate prediction remains fixed at $c_{\mathrm{nom}}$. Under uniform illuminance scaling, the model fully preserves ranking consistency (Spearman~$\rho = 0.641$), since the relative ordering of camera viewpoints remains unchanged. However, the absolute estimation error increases as illuminance decreases. The surrogate remains within the in-distribution error range ($\mathrm{RMSE} < 0.154$) when illuminance is above approximately 46$\%$ of the training condition, beyond which overestimation becomes increasingly significant.

These results indicate that the surrogate model generalizes well under moderate environmental variation and maintains reliable ranking behavior across different viewing conditions. However, its performance degrades under severe occlusion and extreme low-light conditions due to the absence of explicit visibility and lighting representations in the model input. These limitations motivate future work on occlusion-aware and illuminance-conditioned surrogate models to further improve robustness in challenging real-world environments.

The ray-tracing accuracy is also tested, comparing the visibility rate given by our proposed system using a single ray test, giving a score of 1 if the ray hits its target AABB and 0 otherwise. An environment with a perception target (a human) similar to \cref{fig:occlusion_heatmap} is evaluated with a wall, a thin wall (partial occlusion) and a fence-like wall (so target is visible through). The test was evaluated against a YOLO-based boolean detection with a $0.75$ confidence threshold. The single-ray test proved to be a conservative estimator in the fence and thin wall environments, reporting an $90.1\%$ and  $85.1\%$ visibility rate compared to YOLO's $97.0\%$ and $95.0\%$, as the ray is easily intercepted by thin geometric features despite the target being actually visible through gaps. Conversely, it slightly overestimated visibility in solid occlusion ($54.5\%$ estimated vs. $49.0\%$ ground truth detection), likely due specific angles and target-specific inaccuracies, such as a human being more easily distinguishable if the face or silhouette are visible. Overall, the single-ray test approximated the ground truth within a $10\%$ margin, which allowed our use for it as a computationally efficient occlusion estimator.

\subsection{Planning Benchmark Results}
\label{experiment:benchmark_results}
\begin{figure}
    \centering
    \includegraphics[width=0.9\columnwidth]{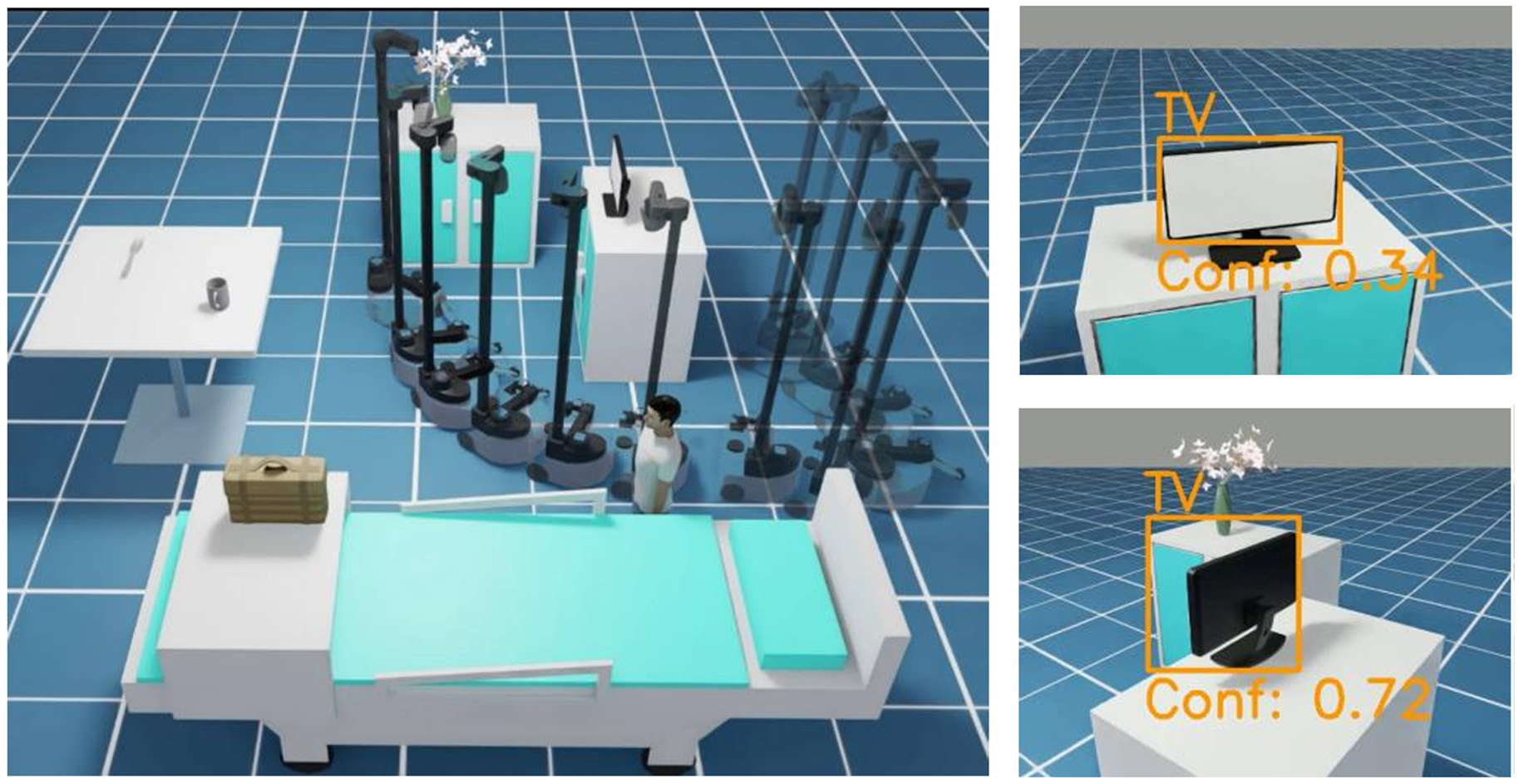}
    \caption{
    Example \textbf{PS-PRM} trajectory for the simulation experiment in \cref{experiment:benchmark_results}. Robot traces and camera images are sampled along the path, fading in as the robot nears the goal. Snapshots show YOLO confidence scores and ground-truth bounding boxes.
    }
    \label{fig:simulation_home}
\end{figure}

For simulation experiments, we create an environment using objects commonly found in a home such as a table and a cabinet (shown in \cref{fig:simulation_home}). Additionally, we introduce five different objects (a cup, a suitcase, a potted plant, a fork, and a TV) common in a home for the robot to detect and track, further mimicking real-world tasks.

We generate $2,500$ motion planning problems by randomly sampling collision-free start and goal configurations for the base, while adjusting the pan and tilt joints so that the robot is directly facing the tracking object. Specifically, we sample 500 problems for each of five objects, with five variations in the object's poses within the environment for each set.

\cref{tab:planning-results} presents results across 2,500 planning problems, evaluating metrics such as planning time, path length, success rate. We use YOLOv9~\cite{wang2024yolov9} and DeepSORT~\cite{Wojke2018deep} for real-time perception performance assessment. It is measured by per-frame object detection rate, average confidence score of object detection, and tracking rate.

For the perception task of object detection and tracking, all methods except \textbf{PS-PRM-frustum}, \textbf{PS-PRM-line}, \textbf{Neural-PS-PRM-frustum}, \textbf{Neural-PS-PRM-line}, and \textbf{RL-Based Active Perception} exhibit detection rates below 50\% and confidence scores under 0.6. This highlights the importance of our method in directly integrating perception scores into the planning process. In contrast, both \textbf{PS-PRM-frustum}, \textbf{PS-PRM-line}, \textbf{Neural-PS-PRM-frustum}, and \textbf{Neural-PS-PRM-line} achieve an object detection rate above 70\%, a confidence score exceeding 0.75, and a tracking rate around 90\%, with \textbf{PS-PRM-line} delivering the best object tracking performance among all methods. 

The \textbf{RL-Based Active Perception} baseline achieves relatively strong perception performance, obtaining a $75.5\%$ detection rate and $80.0\%$ tracking rate, substantially outperforming classical motion-planning baselines such as Post, Rejection, and Manifold-based approaches. However, this improved observability comes at the cost of significantly longer trajectories and reduced planning reliability. In particular, the average path length of \textbf{RL-Based Active Perception} is $21.75$\,rad, which is $44.9\%$ longer than \textbf{PS-PRM-line} and $60.9\%$ longer than \textbf{Neural-PS-PRM-line}. This behavior suggests that the learned policy prioritizes maintaining favorable viewpoints over efficient goal-reaching behavior. Furthermore, the success rate drops to only $57.3\%$, compared to over $95\%$ for all variants of \textbf{PS-PRM} and Neural-\textbf{PS-PRM}, indicating that the learned policy frequently fails to reach the goal region before timeout or becomes trapped in locally suboptimal behaviors. Another limitation of \textbf{RL-Based Active Perception} is the significant training overhead required for each environment and object-monitoring scenario. The method requires approximately $169.8$\,s of training/preprocessing time per scene, whereas the learned surrogate variants Neural-\textbf{PS-PRM-frustum} and Neural-\textbf{PS-PRM-line} require only around $32$\,s, representing approximately a $5.3\times$ reduction in preprocessing time. Although the RL baseline achieves competitive detection performance, its confidence score ($0.634$) remains lower than \textbf{PS-PRM-line} ($0.792$) and Neural-\textbf{PS-PRM-line} ($0.771$). This gap highlights an important distinction between the methods: the RL policy is only indirectly encouraged to maintain observability through reward shaping and a geometric proxy, whereas our approach explicitly optimizes a neural surrogate model trained directly on the downstream perception task.

In contrast, the \textbf{Visibility-Aware Trajectory Optimization} keeps a high reliability and is even the best method for shorter paths ($5.62$rad). However, compared with the PS-PRM and RL methods, its lack of semantic understanding through actual perception information results in a reduced detection rate and confidence. Furthermore, given it uses a simple initialization for the optimization problem (a linear interpolation from start to goal configurations) it has a reduced success rate in comparison to the PS-PRM methods, as the robot might still find itself in collision or be unable to solve for constraints (such as not moving in the y-axis for a non-holonomic robot).

It should be noted that constructing the constrained PRM with perception-based edge scores, using approximately 3,000 nodes across different monitored objects and scenes (in this experiment, 5 different objects and 5 locations each), takes on average 341.5 seconds for \textbf{PS-PRM-line} and 291.7 seconds for \textbf{PS-PRM-frustum}. Once the PRM is built, subsequent planning is very fast, as demonstrated in \cref{tab:planning-results}.

The preprocessing times for \textbf{PS-PRM-frustum} and \textbf{PS-PRM-line} are relatively long due to the high computational cost of rendering the scene to compute perception scores. In contrast, \textbf{Neural-PS-PRM-frustum} and \textbf{Neural-PS-PRM-line} achieve significantly shorter preprocessing times by using a lightweight neural surrogate model to directly estimate perception scores, thereby avoiding costly rendering and computationally intensive perception models. This approach is also advantageous in dynamic scenarios, as perception scores can be efficiently updated when the monitored object or human moves.

Notably, the performance of \textbf{Neural-PS-PRM-frustum} and \textbf{Neural-PS-PRM-line} remains comparable to that of \textbf{PS-PRM-frustum} and \textbf{PS-PRM-line}. Specifically, \textbf{Neural-PS-PRM-frustum} and \textbf{Neural-PS-PRM-line} are approximately ten times faster than the version without the neural surrogate model, while maintaining comparable performance, achieving detection rates within 1\% and confidence scores within 0.01 of the original. Interestingly, the resulting path lengths are also slightly shorter. This suggests that access to ground-truth perception scores may not be essential, and that a well-trained surrogate model can provide sufficiently accurate estimates to achieve a comparable or even better balance between perception quality and motion efficiency.


\cref{fig:simulation_home} illustrates a path generated by \textbf{PS-PRM-line}. demonstrating the observations made at two different stages of the path. Notably, it selects a trajectory that stays slightly farther from the monitored object, resulting in a better viewing angle at the cost of a slightly longer path. We notice that the confidence score is sometimes higher when detecting the back of the screen than the front. This occurs because the back of the monitor is dark and high-contrast relative to the background, whereas the front is lighter and contains glossy or reflective regions. In general, detection confidence produced by a perception model can vary due to multiple factors, including the distribution of training data, color and texture contrast, lighting conditions, surface reflectance, and viewpoint-dependent features. As the learned perception score is trained across a wide range of environments, viewpoints, and object configurations, these artifacts are mitigated and smoothed.
 
These results demonstrate that integrating perception scores into motion planning, especially through a neural surrogate model, enables efficient planning while maintaining high perception quality. This provides a scalable and effective solution for real-time perception-aware robot motion.

\subsection{Human Tracking in a Nursing Setting}
\label{experiment:nuring_experiment}



\begin{figure}
    \centering
    \includegraphics[width=0.9\columnwidth]{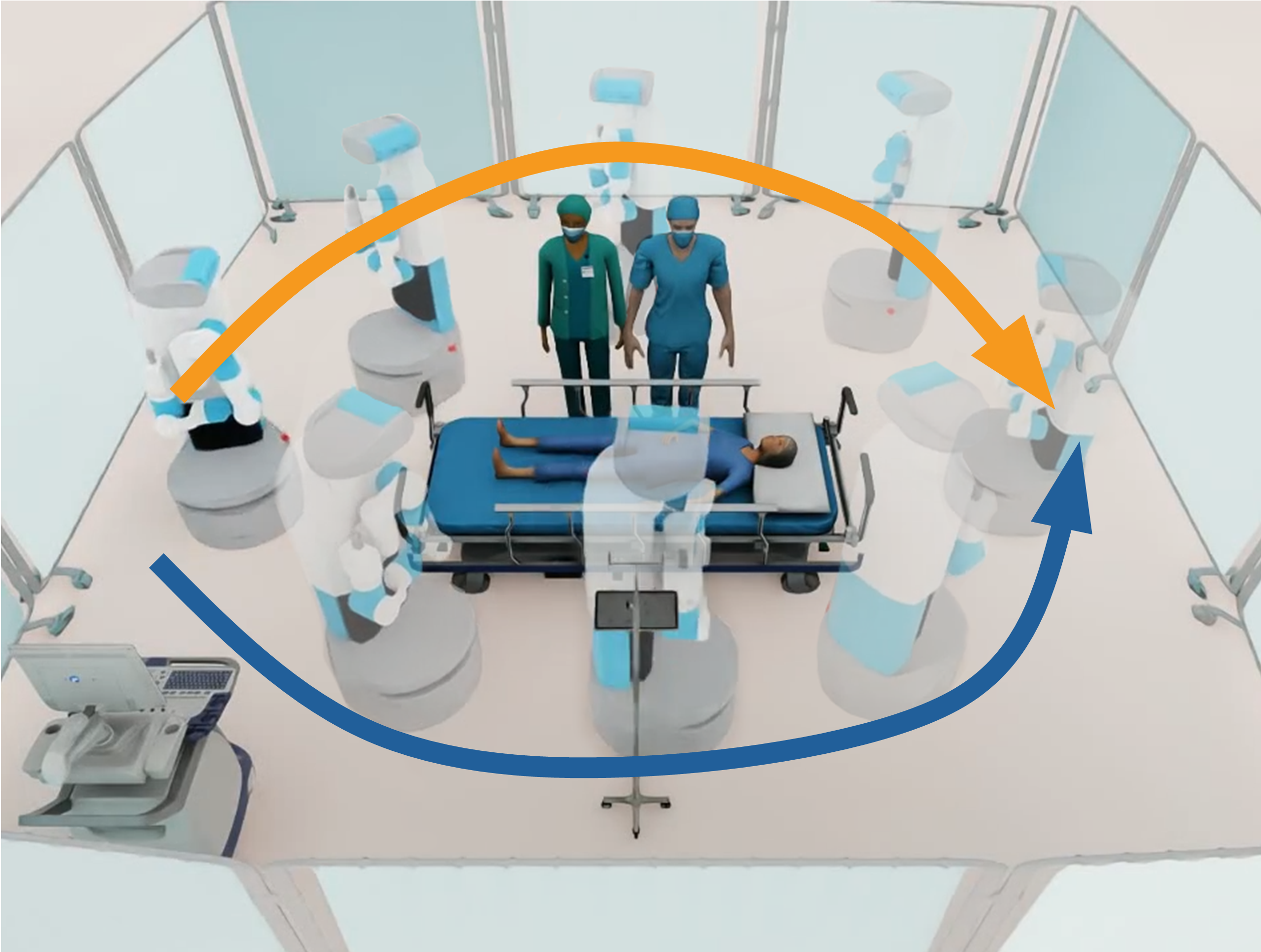}
    \caption{Simulated nursing environment: the robot plans start-to-goal while tracking both masked nurses. \textbf{Manifold-line} takes a shorter path (orange) with poor visibility; \textbf{Neural-PS-PRM-line} takes a longer path (blue) with better detection.}
\label{fig:nursing_environment}
\end{figure}

As shown in \cref{fig:nursing_environment}, we create a nursing environment with two nurses with masks and one patient. The robot is required to plan a path from the start to the goal position while aiming to continuously detect some key features of the nurses. In this experiment, we consider the face detection task, and we use RetinaFace~\cite{deng2019retinaface} for real-time face detection performance assessment. For this multi-person task, we use the midpoint between the two individuals as the monitoring point and the average of their face detection scores to construct the PRM and assess performance. In this experiment, we use the Fetch robot and plan the motion of its base, torso, and camera to perform the assigned task.

For the specific scenario depicted in \cref{fig:nursing_environment}, we evaluate both \textbf{Manifold-line} and \textbf{Neural-PS-PRM-line} over 100 independent runs. Each run of \textbf{Manifold-line} is given a maximum of 60 seconds to compute a solution. For \textbf{Neural-PS-PRM-line}, we construct a PRM and score its nodes and edges by incorporating the face detection confidence score into the planning process with a trained neural surrogate model.

\cref{tab:nursing-results} summarizes the average performance over all runs, reporting metrics such as planning time, path length, face detection rate (for detecting at least one nurse or both nurses), and average face detection confidence score. We observe that \textbf{Manifold-line} tends to choose a shorter path by moving behind the nurses, as shown in \cref{fig:nursing_environment}, which often leads to poor visibility of their faces. In contrast, \textbf{Neural-PS-PRM-line} consistently favors slightly longer and cluttered paths that provide better visibility of the nurses’ faces.

As shown in \cref{tab:nursing-results}, \textbf{Neural-PS-PRM-line} significantly outperforms \textbf{Manifold-line} in both face detection rate and average confidence score. This improvement comes at the cost of increased path length and planning time. Specifically, \textbf{Neural-PS-PRM-line} achieves over 98\% success in detecting at least one nurse and over 63\% for detecting both nurses, while the baseline method fails entirely in the two-face detection task. An example of the routes selected by our method and the baseline is shown in \cref{fig:nursing_environment}.

\subsection{Conflicting target placements}  
\label{experiment:conflicting_targets}

\begin{figure}
    \centering
    \includegraphics[width=0.85\columnwidth]{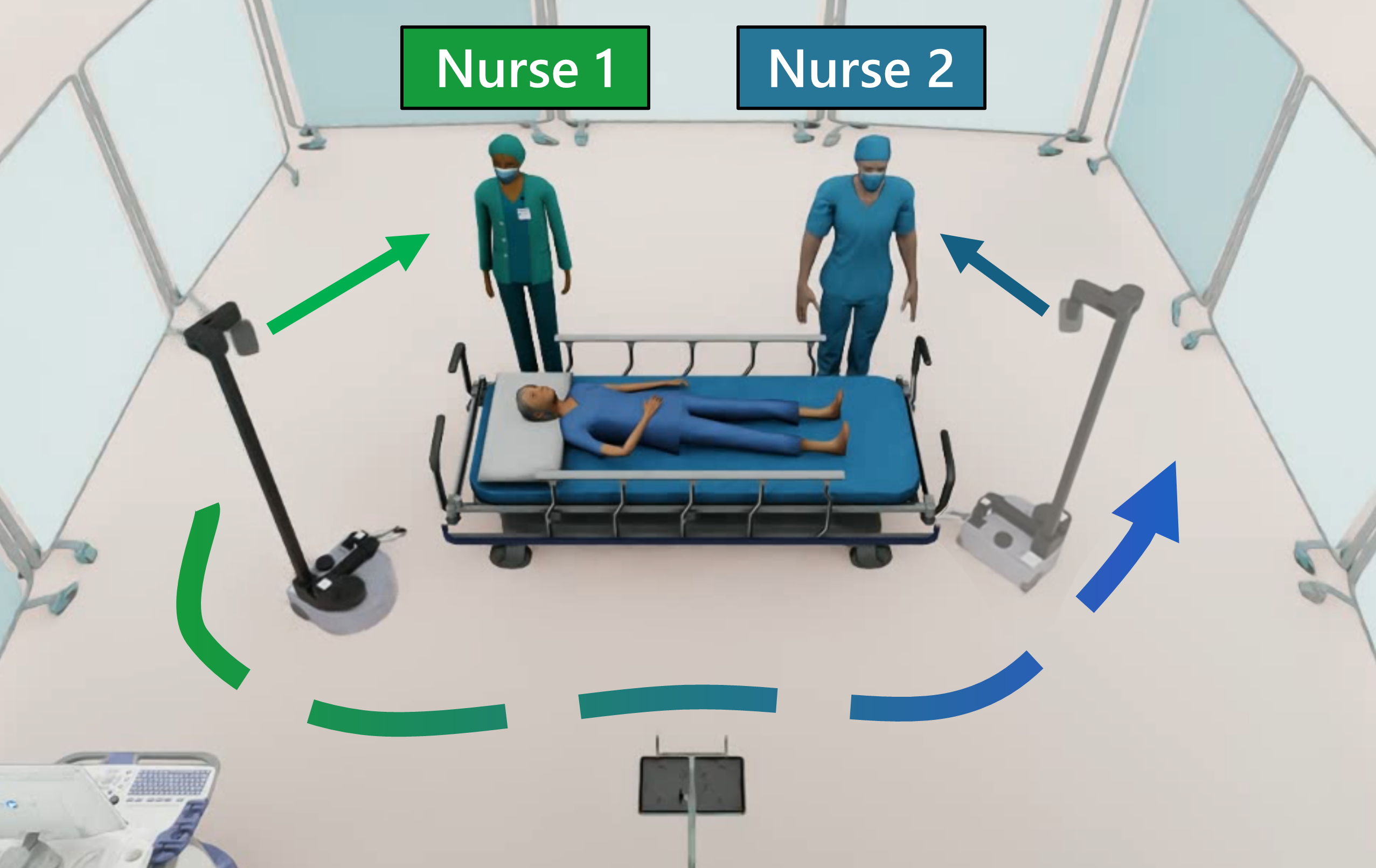}
    \caption{
    Distance-based priority under conflicting perception objectives: with the two nurses spatially separated, the Stretch robot shifts attention from Nurse~1 to Nurse~2 as it moves left to right.
    }
    \label{fig:multi_nurse}
\end{figure}

\cref{fig:multi_nurse} illustrates a scenario with conflicting target placements, where the planner must balance multiple perception objectives. In this experiment, the two nurses are spatially separated such that maintaining high-quality perception of both targets simultaneously is not feasible. To resolve this conflict, the planner does distance-based target-selection. As the Stretch robot moves from left to right, it initially prioritizes tracking Nurse~1, and shifts attention toward Nurse~2 as the robot approaches closer, reflecting the changing importance of the perception objectives. Given the requirement of constraint satisfaction for this problem, the constraint tolerance is increased for this experiment to work, allowing the camera to aim away from one object into another during planning.

\begin{figure}
    \centering
    \includegraphics[width=0.85\columnwidth]{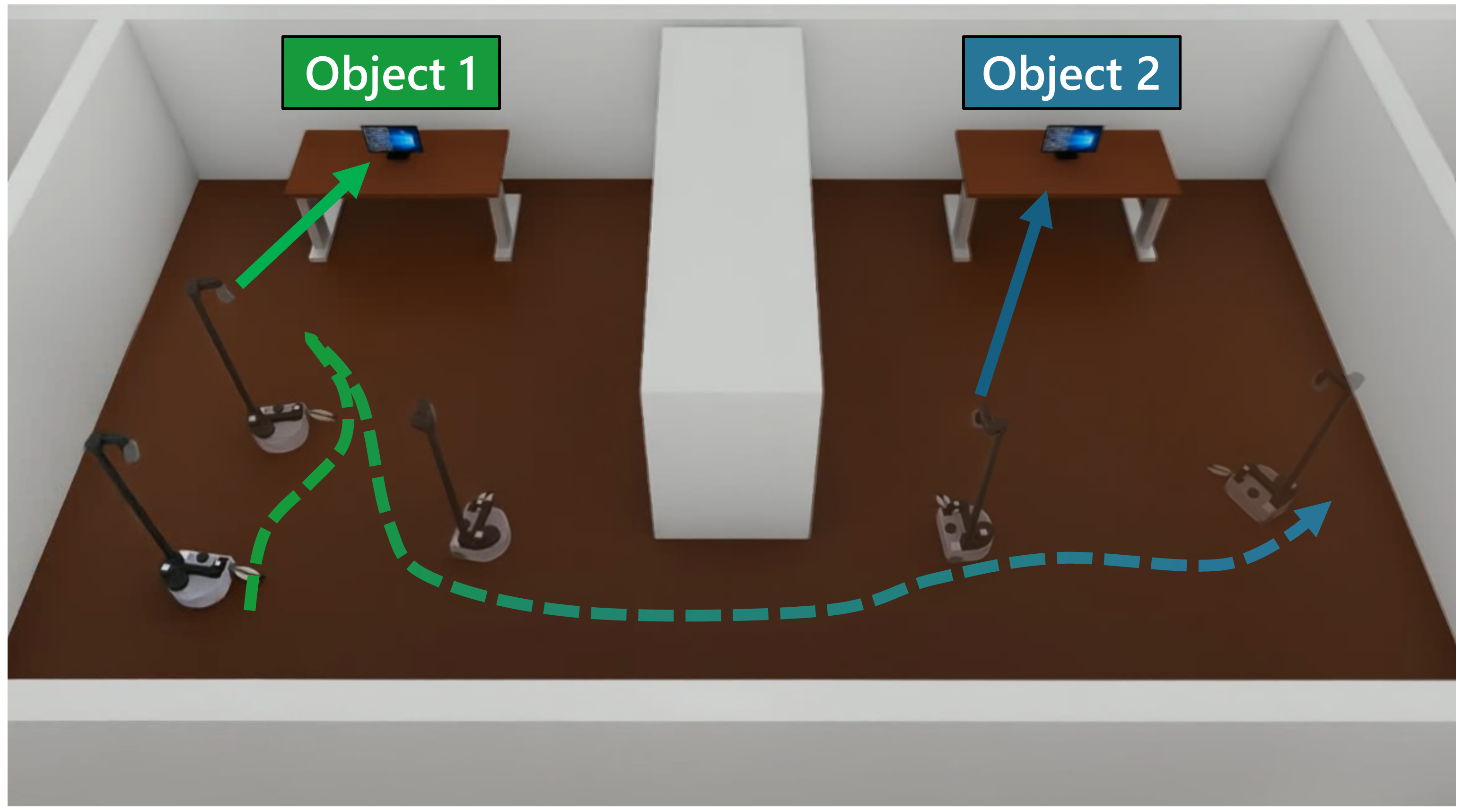}
    \caption{Planning through a narrow passage: distance-based priority shifts which object the camera focuses on as the robot crosses.}
\label{fig:multiroom}
\end{figure}

Similarly, in \cref{fig:multiroom} the conflict arises from the complex collision environment the robot has to move in, with a narrow passage between the start and goal configuration. The distance-based priority metric induces the change in the current tracked object, while the sampling-based planning allows to successfully solve the narrow passage movement.

\begin{table}[t]
\centering
\caption{Comparison of \textbf{Manifold-line} and \textbf{Neural-PS-PRM-line} in a simulated nursing environment. ``Plan. Time'' includes PRM construction for \textbf{Neural-PS-PRM-line}. ``Det. Rate (1/2 face)'' is the detection rate for one or both nurses; ``Conf. Score'' is the average face-detection confidence.}
\label{tab:nursing-results}
\resizebox{\columnwidth}{!}{
\begin{tabular}{lcc|ccc}
\toprule
 & \multicolumn{2}{c}{Motion Planning} & \multicolumn{3}{c}{Human Tracking} \\
\cmidrule(lr){2-3} \cmidrule(lr){4-6} 
 \multirow{2}{*}{Method} & Plan. & Path & Det. Rate & Det. Rate & Conf. \\
 & Time (s) & Len. (rad) & (1 face) & (2 faces) & Score \\ 
\midrule
\rowcolor{gray!1} Manifold-line & 5.31 & 6.89 & 31.53\% & 0.00\% & 0.49 \\ 
\rowcolor{gray!10} Neural-PS-PRM-line  & 32.91 & 11.51 & 98.40\% & 63.29\% & 0.81 \\ 
\bottomrule
\end{tabular}
}
\end{table}

\subsection{Planning in a Dynamic Environment}
\label{experiment:dynamic_environment}

\begin{figure}
    \centering
    \includegraphics[width=0.9\columnwidth]{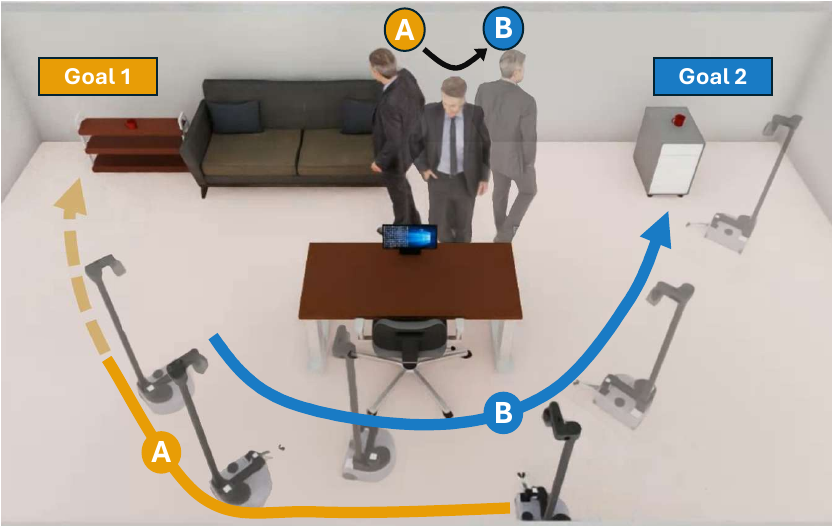}
    \caption{Dynamic planning for human face detection: as the human rotates, the Stretch~2 robot replans from the shelf (Goal~1, orange) to the cabinet (Goal~2, blue) to preserve visibility. \textbf{Neural-PS-PRM-line}'s online replanning achieves much higher detection rates than static planning.}
\label{fig:dynamic_environment}
\end{figure}

As shown in \cref{fig:dynamic_environment}, we construct a dynamic environment containing a human, a cabinet, a shelf, and other common household objects such as a sofa, table, and chair. The robot is tasked with planning a path from its initial position to reach one of two goal locations (the cabinet or the shelf), while continuously detecting key facial features of the human. This experiment focuses on a face detection and tracking task, utilizing RetinaFace~\cite{deng2019retinaface} to assess real-time face detection performance. All trials are conducted using the Stretch 2 robot. During execution, the human moves and rotates from side to side, requiring the robot to dynamically select which goal to pursue in order to maintain high face detection performance.

For the specific scenario illustrated in \cref{fig:dynamic_environment}, we evaluate both \textbf{Manifold-line} and \textbf{Neural-PS-PRM-line} over 100 independent runs. Each \textbf{Manifold-line} trial is given a maximum of 60 seconds to compute a plan. In contrast, \textbf{Neural-PS-PRM-line} constructs a PRM and assigns perception-aware scores using a trained neural surrogate model based on face detection confidence. Unlike \textbf{Manifold-line}, which does not account for the human’s dynamic movements, \textbf{Neural-PS-PRM-line} supports online replanning during execution as described in \cref{sec:replanning}. \Cref{tab:dynamic-results} presents the average results across all runs, evaluating metrics such as planning time, path length, face detection rate, and average confidence score.

As expected, \textbf{Manifold-line} consistently selects a single goal and results in very low face detection performance. In contrast, \textbf{Neural-PS-PRM-line} initially chooses the path toward the shelf on the left, as the human begins facing that direction. During execution, the robot replans in real time and switches to the path toward the cabinet on the right as the human turns, thereby maintaining better visibility of the face. As shown in \cref{tab:dynamic-results}, \textbf{Neural-PS-PRM-line} significantly outperforms \textbf{Manifold-line} in both face detection rate and average confidence score, while also demonstrating successful online replanning. An example of the routes selected by our method is shown in \cref{fig:dynamic_environment}.

In summary, these experiments confirm that our approach \textbf{PS-PRM} is effective for planning simultaneous perception and motion for high-DoF robots in practical human-robot interaction applications in dynamic environments.

\begin{table}[t]
\centering
\caption{Comparison of \textbf{Manifold-line} and \textbf{Neural-PS-PRM-line} in a dynamic, human-aware scenario. ``\#Plans'' is the average number of replans during execution; ``Det. Rate'' and ``Conf. Score'' are face-detection rate and average confidence.}
\label{tab:dynamic-results}
\resizebox{\columnwidth}{!}{
\begin{tabular}{lccc|cc}
\toprule
 & \multicolumn{3}{c}{Motion Planning} & \multicolumn{2}{c}{Face Detection} \\
\cmidrule(lr){2-4} \cmidrule(lr){5-6} 
 Method & Plan. Time (s) & \#Plans & Path Len. (rad) & Det. Rate & Conf. Score \\ 
\midrule
\rowcolor{gray!1} Manifold-line & 0.17 & 1 & 5.48 & 74.13\% & 0.55  \\ 
\rowcolor{gray!10} Neural-PS-PRM-line & 2.49 & 2.12 & 10.53 & 90.53\% & 0.73 \\ 
\bottomrule
\end{tabular}
}
\end{table}

\subsection{Mobile Manipulator Planning in Blind Spot Alley}
\label{experiment:ur5_demo}
In \cref{fig:ur5_demo}, we illustrate another scenario where a mobile manipulator (a UR5 robot mounted on a differential-drive base) moves from a start configuration at the right-side corner to a goal configuration at the left-side corner, navigating through a cluttered environment. During this motion, the robot must maintain high perception performance, measured by its ability to continuously track a monitor placed at the far end of the cross lane. We use \textbf{Neural-PS-PRM-frustum} in this case to plan full-body motion for the robot, ensuring that the monitor remains consistently detected with high confidence scores. As shown in \cref{fig:ur5_demo}, the robot continuously adjusts its arm to maintain a clear view of the monitor with its wrist-mounted camera, while simultaneously navigating its differential-drive base to avoid obstacles. The accompanying camera views confirm that the monitor is successfully detected throughout the entire motion. This capability is broadly applicable, enabling the robot to perceive blind spots in the environment to avoid collisions or continue tracking critical objects or humans.

\subsection{Real Robot Experiments}
\label{experiment:real_robot}
\subsubsection{Static Object Tracking Task}
\begin{figure*}
    \centering
    \includegraphics[width=0.8\textwidth]{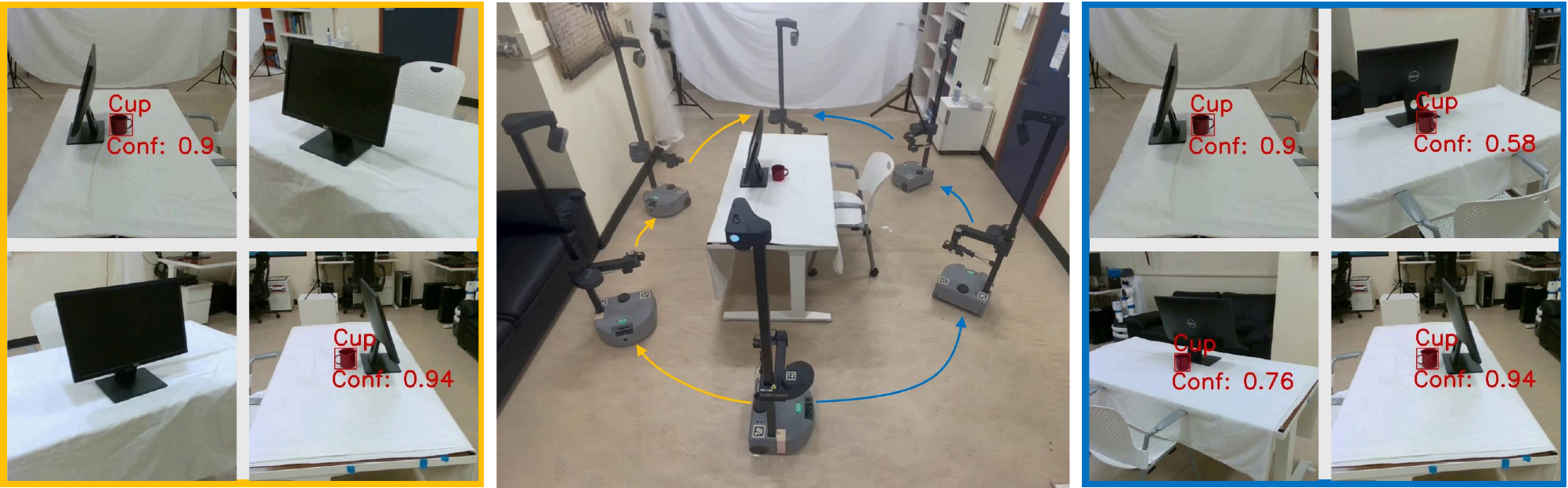}
    \caption{
    Real-robot cup-detection experiment: motion paths for \textbf{Manifold-line} (orange) and \textbf{PS-PRM-line} (blue), with camera views for each shown top-to-bottom, left-to-right.
    }
    \label{fig:real_robot}
\end{figure*}

In this experiment, as shown in \cref{fig:real_robot}, we set up an environment where the robot must choose a path, either left or right around the table, to reach the other side. Simultaneously, the robot is tasked with continuously detecting the cup placed on the table. The left path is shorter, but visibility of the cup is poor due to occlusion by the monitor. The right path is longer, as it is partially blocked by the chair, but offers better visibility of the cup.

We compare \textbf{PS-PRM-line} (blue path) with \textbf{Manifold-line} (orange path). As illustrated in \cref{fig:real_robot}, \textbf{PS-PRM-line} takes a longer route to avoid occlusion by the monitor, ensuring continuous detection of the cup, while the \textbf{Manifold-line} opts for the shorter path, which leads to the cup being occluded. This is further supported by the robot's camera view images (displayed sequentially from top left to bottom right), showing that \textbf{PS-PRM-line} detects the cup in significantly more frames compared to \textbf{Manifold-line}.

\subsubsection{Dynamic Human Tracking Task}
\begin{figure}
    \centering
    \includegraphics[width=0.9\columnwidth]{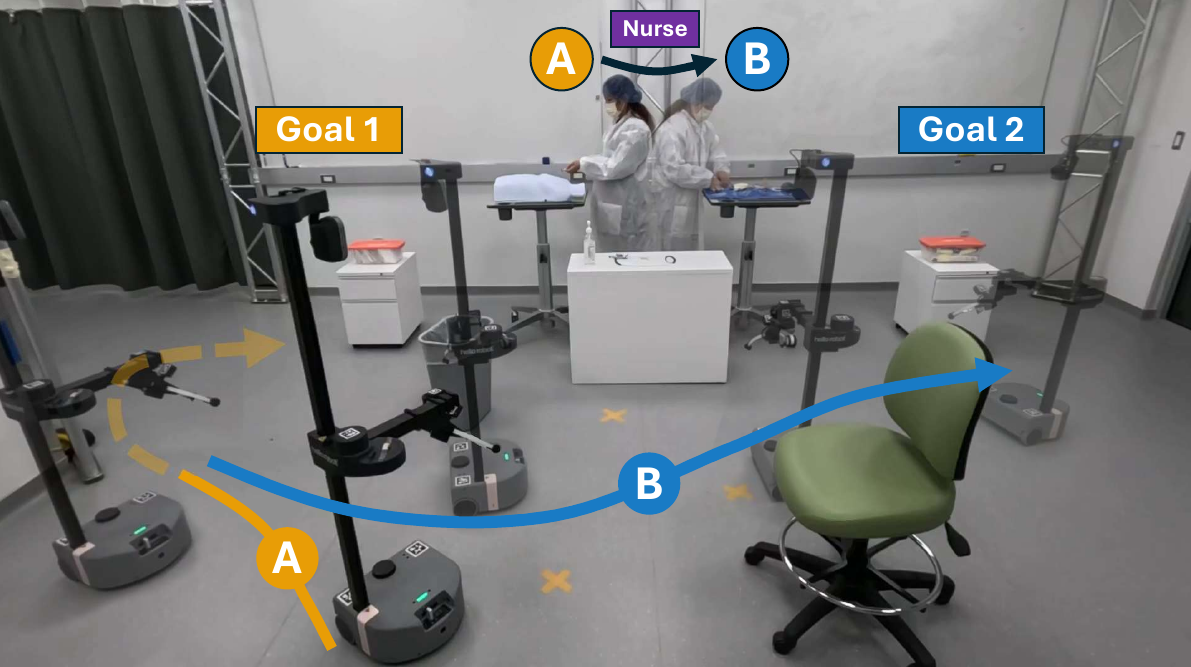}
    \caption{
Real-robot face-detection experiment: as the nurse rotates, the Stretch~2 robot replans from the left cabinet (Goal~1, orange) to the right cabinet (Goal~2, blue) to preserve visibility, using \textbf{Neural-PS-PRM-line}'s online replanning.
    }
    \label{fig:real_robot_dynamic}
\end{figure}

In this experiment, illustrated in \cref{fig:real_robot_dynamic}, we demonstrate the deployment of our system in a realistic nurse training scenario. The environment includes a nurse engaged in her workspace, surrounded by typical elements such as work tables, a chair, a platform, a garbage can, and two cabinets that serve as potential goal locations where the robot needs to grab some tools for the nurse. The task requires the Stretch 2 robot to navigate to one of these locations while maintaining reliable face detection of the nurse. As the nurse rotates during the task, the robot must adapt to changes in facial visibility. Our system initially plans a path to the left cabinet (Goal 1) when the nurse faces left, but dynamically replans to the right cabinet (Goal 2) as she turns right, preserving optimal visibility. This experiment highlights how \textbf{Neural-PS-PRM-line} enables evaluating trajectories on real-time and replanning, a critical requirement in nurse training and collaborative medical environments.

\subsubsection{Human Tracking with Moving Obstacle and Goal}
\begin{figure}[H]
    \centering
    \includegraphics[width=0.86\columnwidth]{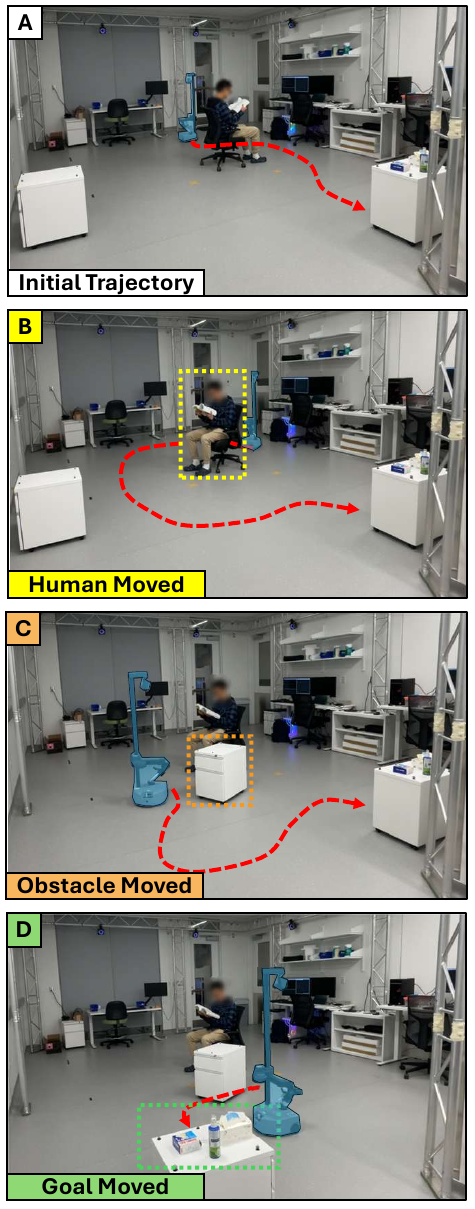}
    \caption{Dynamic human, obstacle, and goal movement: the Stretch~2 robot (blue) switches paths as the person rotates (B), replans around a moved obstacle (C), and replans again as the goal cabinet moves (D).
    }
    \label{fig:moving_obstacle}
\end{figure}

In this experiment (illustrated in \cref{fig:moving_obstacle}), we demonstrate our system’s performance in a realistic, dynamic scenario. The environment contains a seated person reading a book, surrounded by two cabinets that serve respectively as an obstacle and a goal location the robot must reach. The task requires the Stretch 2 robot to reach a configuration near the goal cabinet while maintaining reliable face tracking of the person. As the person naturally rotates during the task, the robot must adapt to changes in their pose and the obstacle locations. When the person initially faces right, the robot selects a path on the right side; as the person turns left, the system dynamically replans and shifts to a left-side path to preserve optimal perception. We then introduce additional environmental changes: the obstacle cabinet is moved mid-execution, forcing the robot to replan in real time to avoid collision, and we further demonstrate that the robot can plan toward a moving goal as the goal cabinet changes position. For experiment setup, we use an OptiTrack motion capture system to track the positions and orientations of both the cabinets and the human; this could be replaced with alternative pose tracking solutions.

By exploiting the biased sampler and triggering replanning via movement thresholds (determined by collision paddings), the planner achieves execution rates of up to 2Hz. It also generated an average of 8 trajectories per run, more than the 3 environment changes. This highlights the capability to plan multiple times on single scene alterations. Overall, this experiment shows how Neural-PS-PRM-line enables collision- and perception-aware online replanning, supporting continuous coordination between motion and perception in the presence of moving obstacles and shifting perception target and motion goal locations.

\subsubsection{Manipulator Grasping Task with Human Tracking Constraints}
\begin{figure}
    \centering
    \includegraphics[width=0.85\columnwidth]{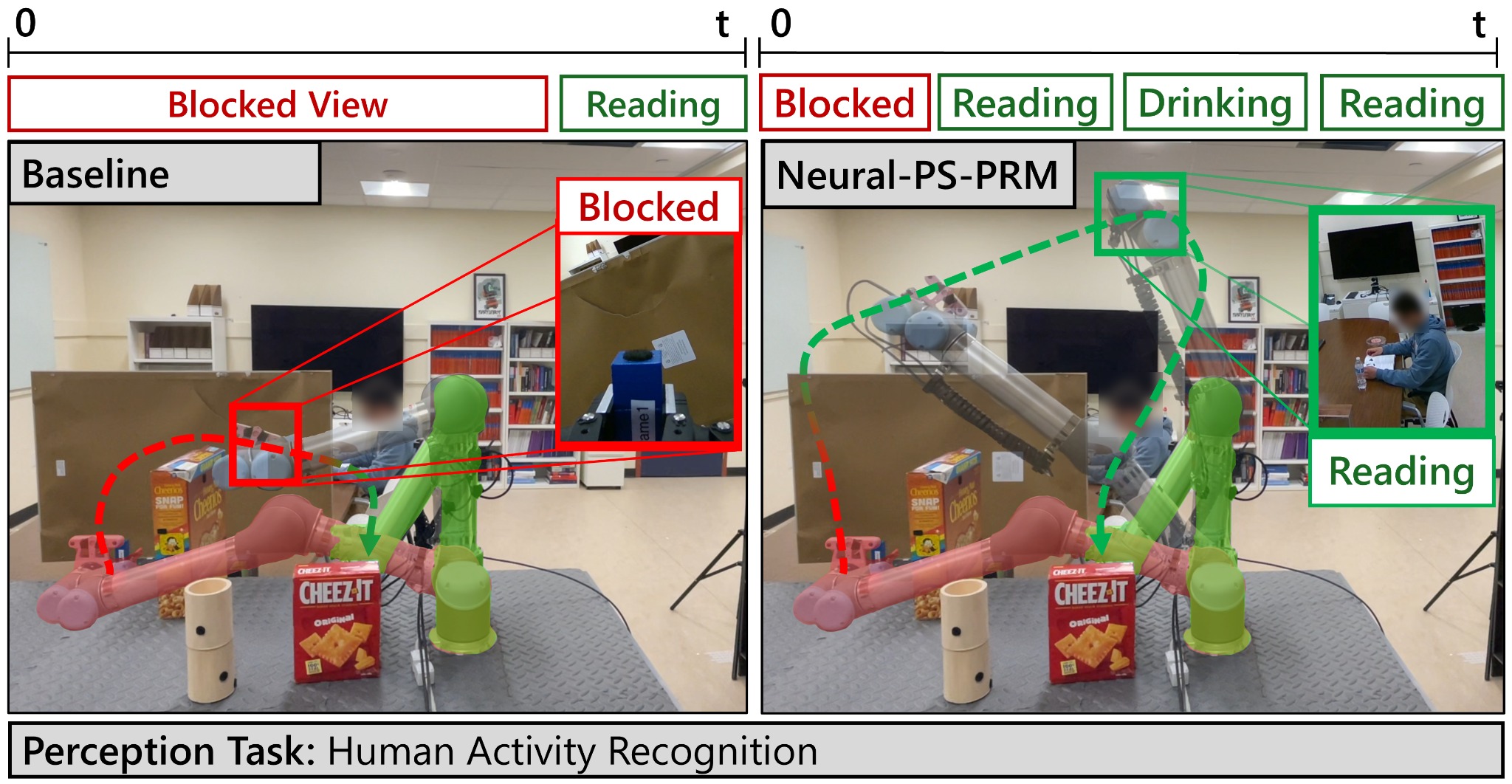}
    \caption{
    Perception-constrained planning for human activity recognition on a 6-DoF UR5. Without perception costs (short path), the person is largely occluded; the \textbf{PS-PRM} trajectory detours to maintain visibility. Trajectories are dotted, red when occluded and green when visible.
    }
    \label{fig:real_robot_ur5}
\end{figure}

In this experiment (illustrated in \cref{fig:real_robot_ur5}), we demonstrate our system’s performance on a real-robot grasping task using a UR5 manipulator. The environment contains a seated person reading a book, and the robot is required to grasp an object while continuously monitoring the person’s activity. A wooden board introduces additional occlusion that can partially block the wrist-mounted camera view. The robot’s task is to move from left to right while maintaining human activity detection for as long as possible.

During the execution, the person alternates between reading a book, picking up a water bottle to drink, and then continuing to read. Human activity recognition is performed using a vision–language–based activity detection model~\cite{kwon2023efficient}. As shown in the left subfigure, the baseline method \textbf{PS-PRM-line}, which does not incorporate occlusion reasoning, selects the shortest path to the goal. However, this trajectory fails to observe the person’s activity for most of the execution and, in particular, misses the drinking action. In contrast, as shown in the right subfigure, \textbf{Neural-PS-PRM-line} plans a longer trajectory that moves above the wooden board, maintaining visibility of the person throughout most of the motion. As a result, it successfully captures the drinking activity and accurately detects the time intervals during which the person is reading.


This real-robot experiment demonstrates the practical applicability of our method to manipulation tasks that require human awareness under occlusion.

In summary, these real-robot experiments show that \textbf{PS-PRM} is suitable for tasks requiring simultaneous perception and motion in dynamic environments.

\section{Discussion}
\label{sec:conclusion}
In this work, we address the problem of planning robot motions for a high-degree-of-freedom (\dof) robot that effectively achieves a given perception task while the robot and the perception target move in a dynamic environment. We propose a GPU-parallelized perception-score-guided probabilistic roadmap planner with a neural surrogate model (\psprm): a roadmap-based planner that explicitly incorporates the estimated quality of a perception task into motion planning for high-\dof robots. Our method uses a learned model to approximate perception scores and leverages GPU parallelism to enable efficient online replanning in dynamic settings.

Our extensive experiments across multiple platforms, including the UR5 robot, the Fetch robot, and Hello Robot's Stretch 2 robot, demonstrate the consistent benefits of PS-PRM over conventional planning pipelines. In a range of realistic environments, including cluttered home settings, blind-spot navigation, and dynamic human-robot interaction scenarios, PS-PRM significantly improves perception outcomes while maintaining or modestly increasing motion costs. For instance, we observe at least a 1.6$\times$ improvement in object detection rates in home settings and over a 2.1$\times$ improvement in face detection rates in crowded nursing environments. Moreover, the neural surrogate model enables planning that is approximately 10$\times$ faster than approaches relying on ground-truth perception pipelines, while achieving comparable detection accuracy and sometimes even shorter paths.

Our case studies highlight the practical value of our method in diverse perception-aware planning tasks: in the UR5 scenario, the arm continuously reorients to maintain a clear view of a monitor despite occlusions; in the Fetch robot experiments, the planner actively selects longer but more perceptually informative paths through nurse-populated corridors; and in dynamic human-robot scenarios with the Stretch robot, the system successfully replans on-the-fly to maintain face visibility despite human movement and rotation.

While our work focuses on object and face detection, the underlying framework is general and applicable to other perception-driven tasks such as semantic mapping, affordance detection, or visual servoing. PS-PRM serves as a bridge between motion generation and high-level perception, enabling robots to make informed decisions that account for both geometry and visibility constraints.

This direction opens several promising avenues for future research. First, while the neural surrogate model significantly reduces computation time and is robot- and environment-agnostic, it is currently trained offline using a fixed set of perception tasks, object categories, and environmental conditions. Future work could explore domain adaptation, continual learning, or occlusion- and illuminance-aware surrogate models to improve robustness under out-of-distribution scenarios and varying real-world environments. Second, incorporating uncertainty estimates from the surrogate model may further enhance robustness, particularly in ambiguous or noisy scenes. Third, there remains opportunities for further GPU parallelization, such as incorporating ideas from~\cite{akinc2005probabilistic} to further accelerate PRM construction and query.
An interesting extension of this work is to explore how the framework can be adapted to a broader range of perception tasks and application domains, such as semantic segmentation or 3D reconstruction.

Finally, real-world deployments in assistive and collaborative environments, such as smart homes and healthcare, will require consideration of social and ethical factors. For example, in nursing applications, robots may need to detect faces, interpret gestures, and maintain situational awareness while respecting privacy. Although this work is focused on training simulations, a natural extension is to integrate privacy-preserving constraints into the planning process~\cite{shome2023privacy}, allowing robots to intelligently navigate and perceive while adhering to user-centered constraints. 

In summary, this work demonstrates the effectiveness of perception-guided planning for high-DoF robots in both static and dynamic environments. By jointly reasoning about motion and perception, PS-PRM offers a flexible and generalizable foundation for building more capable, context-aware robotic systems.

\printbibliography{}

\begin{IEEEbiography}[{\includegraphics[width=1in,height=1.25in,clip,keepaspectratio]{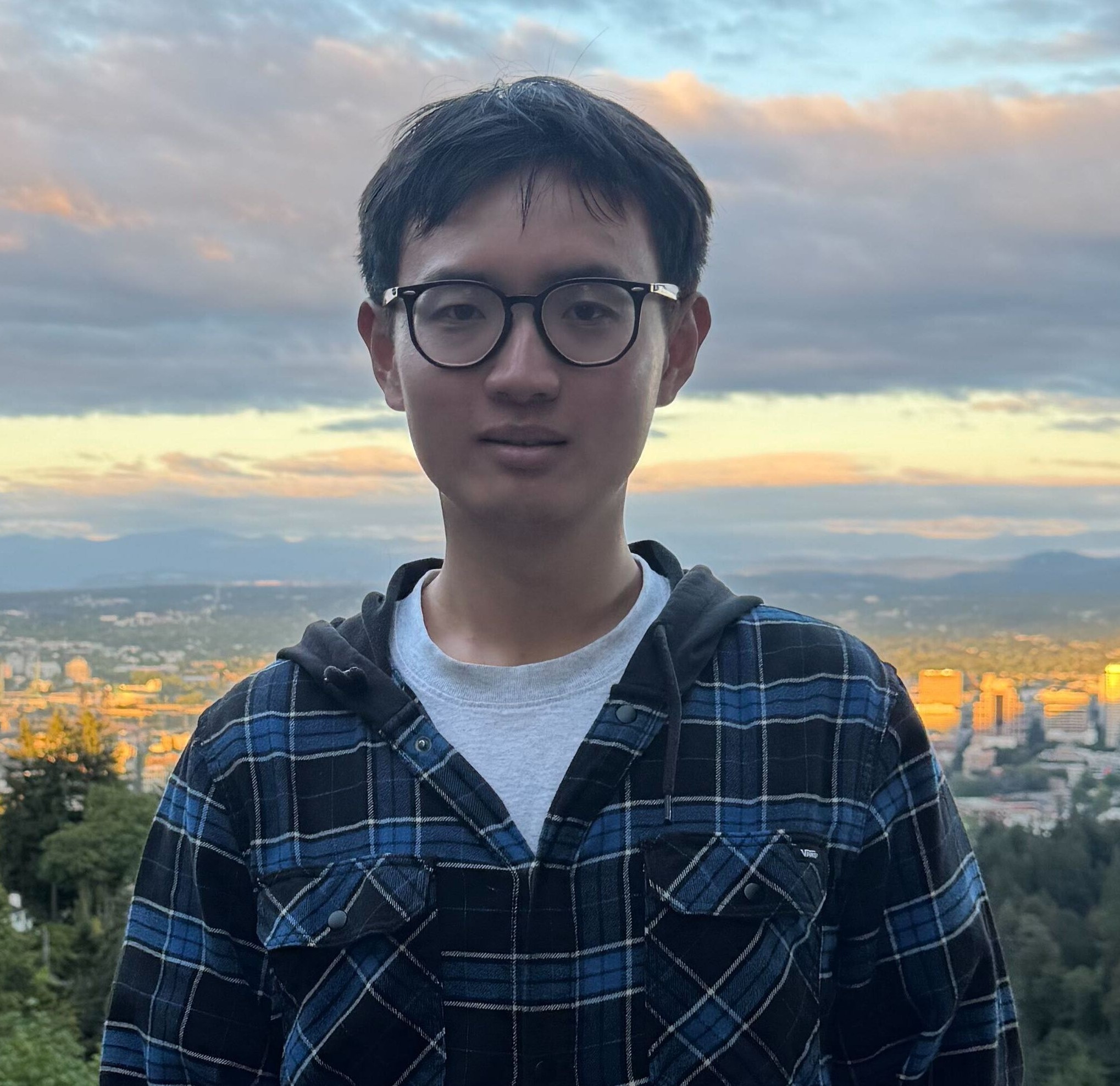}}]{Qingxi Meng}
received the B.S. degree in electrical engineering from the University of Illinois at Urbana-Champaign, Urbana, IL, USA, in 2020, and the M.S. degree in electrical engineering from Stanford University, Stanford, CA, USA, in 2022. He is currently working toward the Ph.D. degree in computer science with Rice University, Houston, TX, USA.

His research interests include perception-aware motion planning, GPU-accelerated planning algorithms, and robot perception for human-robot interaction in healthcare environments.
\end{IEEEbiography}

\begin{IEEEbiography}[{\includegraphics[width=1in,height=1.25in,clip,keepaspectratio]{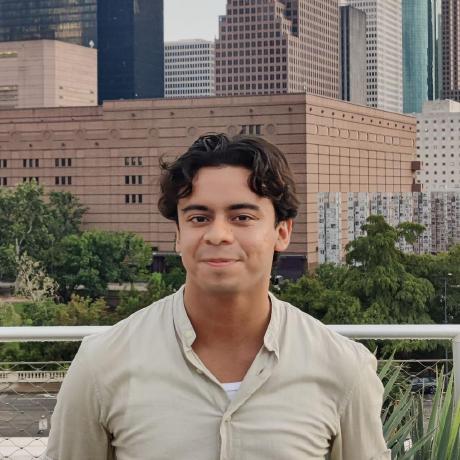}}]{Emiliano Flores}
received the B.Sc. degree in robotics engineering from Tecnológico de Monterrey, Mexico, in 2025. He is currently working toward the Ph.D. degree in computer science with Rice University, Houston, TX, USA, under the supervision of Prof. Lydia E. Kavraki.

His research interests include hardware-accelerated, perception- and task-aware motion planning, optimization-based methods and long-horizon planning. 
\end{IEEEbiography}

\begin{IEEEbiography}[{\includegraphics[width=1in,height=1.25in,clip,keepaspectratio]{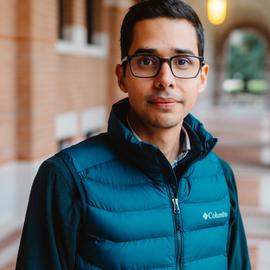}}]{Carlos Quintero-Peña}
received the B.S. degree in electronic engineering and the M.Sc. degree in electronic and computer engineering from Universidad de los Andes, Bogotá, Colombia, in 2009 and 2010, respectively, and the Ph.D. degree in computer science from Rice University, Houston, TX, USA, in 2024, under the supervision of Prof. Lydia E. Kavraki. He was a recipient of a Fulbright Scholarship in 2019.

His research interests include motion planning for high-degree-of-freedom robots, optimization-based planning under sensing uncertainty, and human-in-the-loop robot planning.
\end{IEEEbiography}

\begin{IEEEbiography}[{\includegraphics[width=1in,height=1.25in,clip,keepaspectratio]{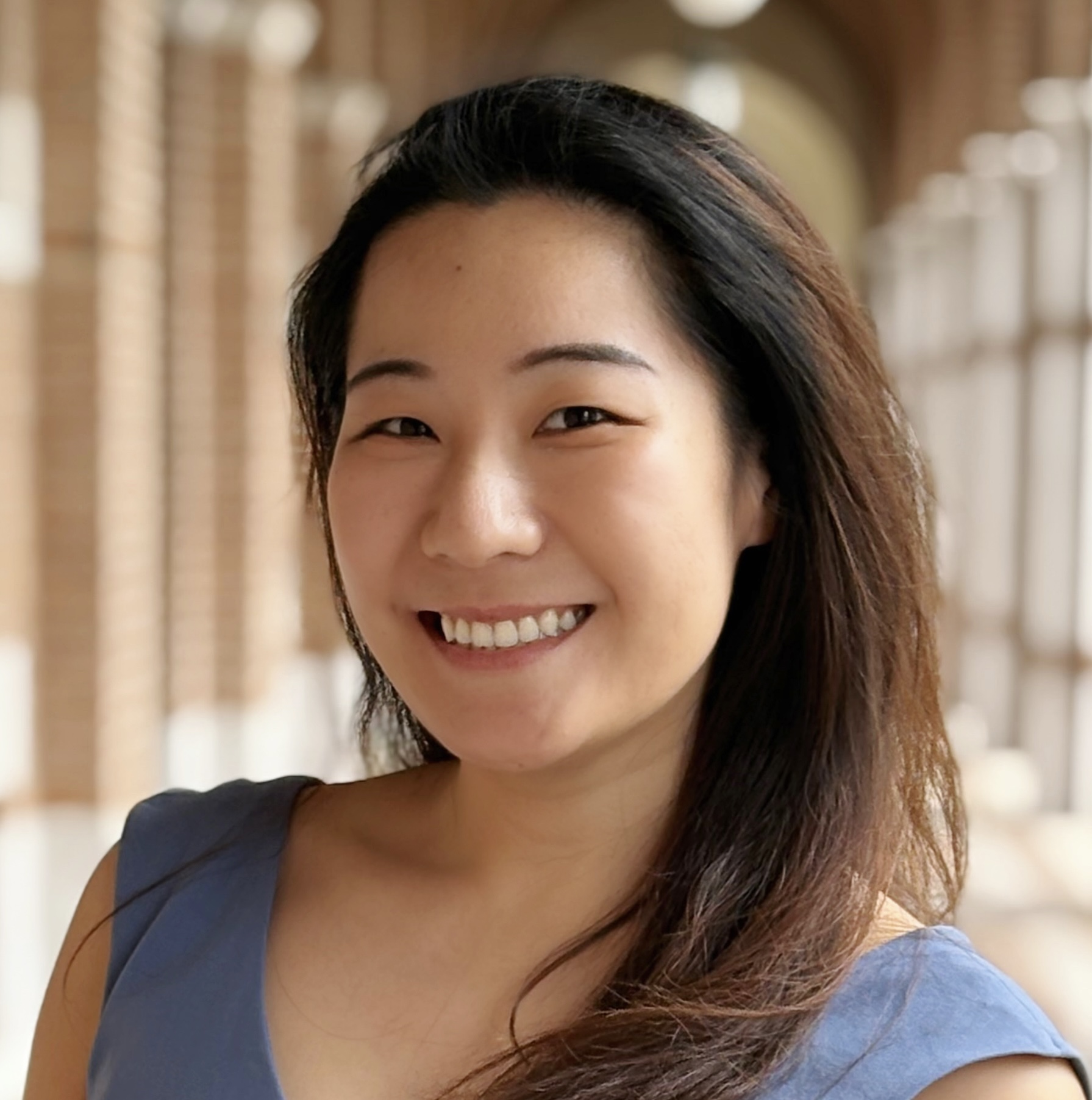}}]{Peizhu Qian}
received the B.S. degree in computer science and mathematics from Simmons College, Boston, MA, USA, in 2019, and the Ph.D. degree in computer science from Rice University, Houston, TX, USA, in 2025, under the supervision of Prof. Vaibhav Unhelkar.

She is currently an Assistant Professor of computer science with the University of Houston, Houston, TX, USA. Her research interests include human-robot interaction, human-centered AI, explainable AI, and healthcare robotics, including intelligent robotic tutors for clinical training.
\end{IEEEbiography}

\begin{IEEEbiography}[{\includegraphics[width=1in,height=1.25in,clip,keepaspectratio]{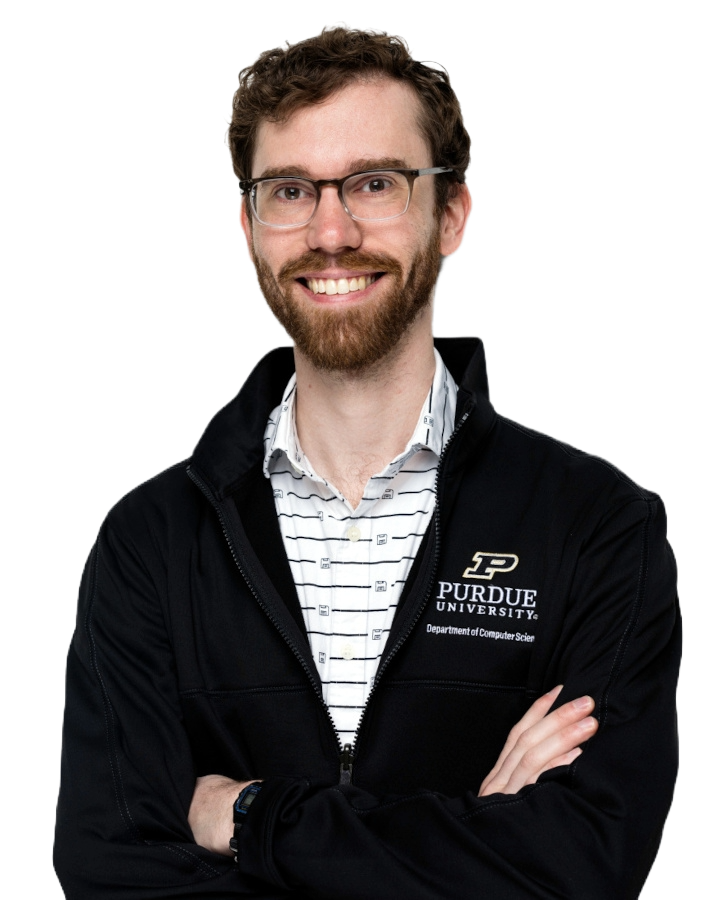}}]{Zachary Kingston}
received the B.S., M.S., and Ph.D. degrees in computer science from Rice University, Houston, TX, USA, in 2016, 2017, and 2021, respectively, under the supervision of Prof. Lydia E. Kavraki.

From 2021 to 2024, he was a Postdoctoral Researcher and Lab Manager with the Kavraki Lab at Rice University. He is currently an Assistant Professor of computer science with Purdue University, West Lafayette, IN, USA, where he leads the Computational Motion, Manipulation, and Autonomy (CoMMA) Lab. His research interests include sampling-based and manifold-constrained motion planning, GPU-accelerated planning, and planning for high-dimensional and soft robotic systems.
\end{IEEEbiography}

\begin{IEEEbiography}[{\includegraphics[width=1in,height=1.25in,clip,keepaspectratio]{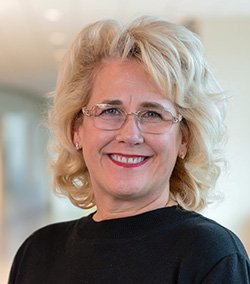}}]{Shannan K. Hamlin}
received the B.S.N. degree from Houston Christian University, Houston, TX, USA, and the M.S.N. and Ph.D. degrees from the University of Texas Health Science Center at Houston, Houston, TX, USA.

She is an Associate Professor of Nursing with the Houston Methodist Academic Institute, where she also serves as Chair of Nursing and Director of the Center for Nursing Research, Education and Practice. She is a board-certified Adult-Gerontology Acute Care Nurse Practitioner and a Fellow of the American College of Critical Care Medicine. Her research interests include hemodynamics and hemorheology in critically ill patients.
\end{IEEEbiography}

\begin{IEEEbiography}[{\includegraphics[width=1in,height=1.25in,clip,keepaspectratio]{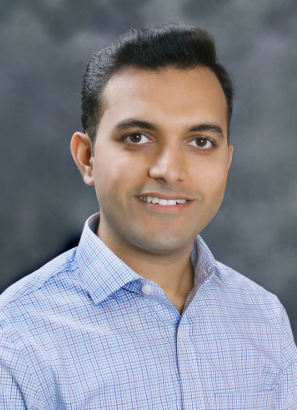}}]{Vaibhav Unhelkar} 
received the Ph.D. degree in autonomous systems from the Massachusetts Institute of Technology, Cambridge, MA, USA, in 2020.

He is an Assistant Professor of Computer Science at Rice University, Houston, TX, USA, where he directs the Teaming Lab. His research lies at the intersection of human modeling, explainable AI, and human-robot interaction, aiming to advance human-machine teaming. In collaboration with domain experts, he has developed robotic assistants for assembly and clinical environments, which work with and alongside humans.

\end{IEEEbiography}

\begin{IEEEbiography}[{\includegraphics[width=1in,height=1.25in,clip,keepaspectratio]{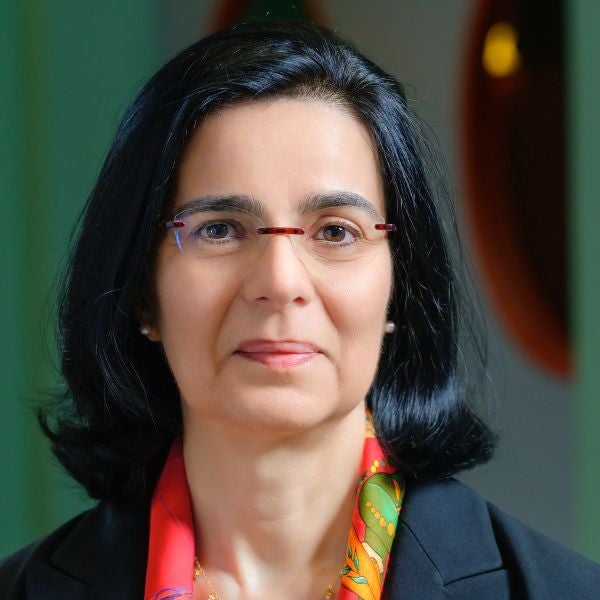}}]{Lydia E. Kavraki}
(Fellow, IEEE) received the Ph.D. degree in computer science from Stanford University, Stanford, CA, USA, in 1995.

She is the Kenneth and Audrey Kennedy University Professor of Computing with Rice University, Houston, TX, USA, and Director of the Ken Kennedy Institute. She coauthored \textit{Principles of Robot Motion} (MIT Press, 2005), pioneered sampling-based motion planning, and leads development of the Open Motion Planning Library. Her research interests include motion planning, human-robot collaboration, and robotics/AI methods for computational biomedicine.

Dr. Kavraki is a Fellow of AAAI, AIMBE, ACM, and AAAS, and a Member of the National Academies of Engineering, Sciences, and Medicine.
\end{IEEEbiography}

\end{document}